\documentclass[10pt,twocolumn,letterpaper]{article}

\usepackage{iccv}
\usepackage{times}
\usepackage{epsfig}
\usepackage{graphicx}
\usepackage{amsmath}
\usepackage{amssymb}
\usepackage{comment}

\usepackage[pagebackref=true,breaklinks=true,letterpaper=true,colorlinks,bookmarks=false]{hyperref}

\iccvfinalcopy 

\def\iccvPaperID{7980} 

\ificcvfinal\fi

\usepackage{placeins}
\usepackage{multirow, rotating}
\newcommand{\Stack}[3][]{#1\begin{tabular}[t]{@{}#2@{}}#3\end{tabular}}
\newcommand{\StackC}[2][]{\Stack[#1]{c}{#2}}

\newcommand{\vect}[1]{\boldsymbol{\mathbf{#1}}}
\usepackage{relsize}
\usepackage{enumitem}

\newcommand{\usik}{PX-CNN-PS}
\newcommand{\usbest}{PX-NET}

\begin{document}

\title{PX-NET: Simple and Efficient Pixel-Wise Training \\ of Photometric Stereo Networks}

\author{Fotios Logothetis$^1$~~~~~~ Ignas Budvytis$^2$ \\
$^1$~ Cambridge Research Laboratory,\\
Toshiba Europe, Cambridge, UK\\
{\tt\small fotios.logothetis@crl.toshiba.co.uk}
 \and
 Roberto Mecca$^{1,2}$~~~~~~  Roberto Cipolla$^2$\\
 $^2$~University of Cambridge\\
   Cambridge, UK\\
 {\tt\small \{ib255,rm822,rc10001\}@cam.ac.uk}
}

\maketitle
\ificcvfinal\thispagestyle{empty}\fi

%
%

\begin{abstract}
\vspace{-0.1cm}
Retrieving accurate 3D reconstructions of objects from the way they reflect light is a very challenging task in computer vision. Despite more than four decades since the definition of the Photometric Stereo problem, most of the literature has had limited success when global illumination effects such as cast shadows, self-reflections and ambient light come into play, especially for specular surfaces.

Recent approaches have leveraged the power of deep learning in conjunction with computer graphics in order to cope with the need of a vast number of training data  in order to invert the image irradiance equation and retrieve the geometry of the object. However, rendering global illumination effects is a slow process which can limit the amount of training data that can be generated.

In this work we propose a novel pixel-wise training procedure for normal prediction by replacing the training data (observation maps) of globally rendered images with independent per-pixel generated data.  We show that global physical effects can be approximated on the observation map domain and this simplifies and speeds up the data creation procedure. 

Our network, PX-NET, achieves the state-of-the-art performance compared to other pixelwise methods on synthetic datasets, as well as the Diligent real dataset on both dense and sparse light settings.

\end{abstract}


\vspace{-0.2cm}
\section{Introduction}
\label{sec:introduction}
Photometric Stereo (PS) is a classical problem in computer vision since the early '80s \cite{Woodham1980}.  PS assumes multiple images from the same viewpoint along with varied illumination and calculates local geometrical features (e.g. normal or depth) at each pixel by exploiting the relation between surface orientation and intensity of reflected light. This is essentially an inverse rendering problem requiring at least three input images in order to have a unique solution. 

Most of the difficulty in retrieving the 3D shape from the light reflected off the object is due to the type of reflection and its non-linear dependence on material properties.  
This is mathematically expressed though the surface bidirectional reflectance distribution function (BRDF) which is determined by the material of the object. 
Over the last forty  years a very wide spectrum of BRDF equations have been proposed to model the light reflection phenomena. Starting from the basic linear light response for diffuse reflection \cite{lambert,Horn1}, more specular behaviour of reflected light have been proposed \cite{Phong75,Blinn:1977,Cook82,lafortune1997non,TorranceSparrow1967,Ward92}. Comparison among numerous BRDFs can be found in \cite{WatsonR01,HavranFM16,NganDM05,NganDM04}. Finally, the recently proposed Disney BRDF \cite{burley2012physically,HillMBCFIHJNPP15} was invented to unify most physical reflection effects including gloss reflection, subsurface scattering and metallic/specular roughness into a unified formulation. 



\begin{figure}[t]
\begin{center} \hspace{-0.4cm}
 \includegraphics[height=0.155\textwidth,trim={6.5cm 5.5cm 7.0cm  4.5cm},clip]{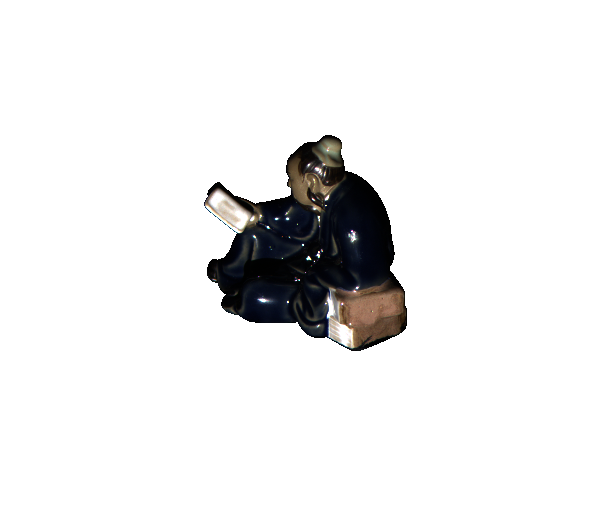} \hspace{-0.25cm}
 \includegraphics[height=0.155\textwidth,trim={6.5cm 5.5cm 7.0cm 
 4.5cm},clip]{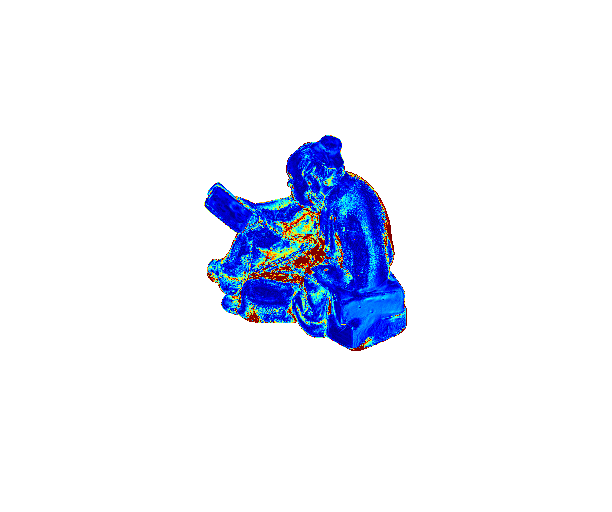} \hspace{-0.25cm}
 \includegraphics[height=0.155\textwidth,trim={6.5cm 5.5cm 7.0cm  4.5cm},clip]{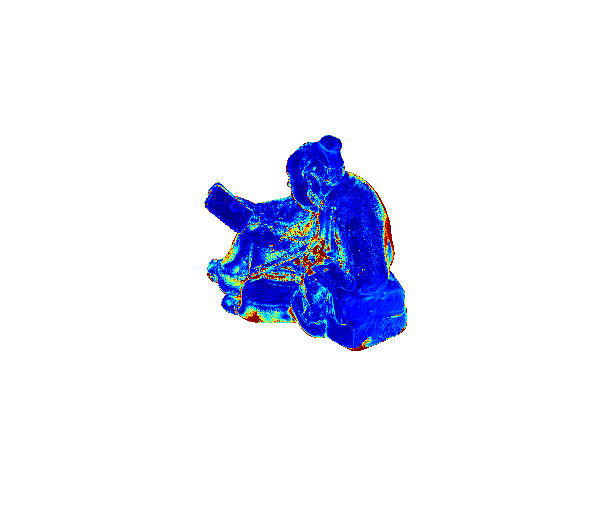} 
 \includegraphics[height=0.155\textwidth,trim={12.5cm 1.0cm 1.0cm  0.5cm},clip]{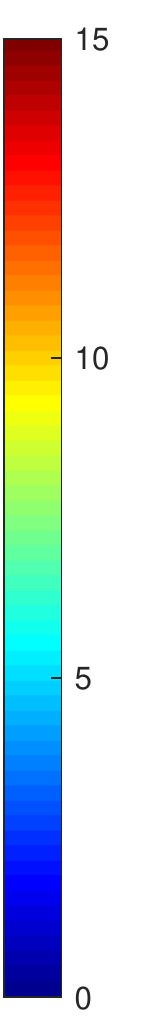} \\
\footnotesize ~~ \hspace{0.02\textwidth}`Reading'  \hspace{0.035\textwidth} \cite{ikehata2018cnn} MAE=$12.6^o$ \hspace{0.4cm}  Proposed MAE=$9.78^o$  \\
 \vspace{0.1cm}
   \caption{Comparison of the proposed approach  versus  \cite{ikehata2018cnn} on `Reading' of the DiLiGenT real benchmark \cite{diligentshi2016benchmark}. The evaluation metric is the mean angular error (MAE) of the computed normal map compared with the ground truth.} \vspace{-0.5cm}
 \label{fig:intro}
\end{center}
 \end{figure}

The above mentioned advancements in computer graphics have enabled convolutional neural network (CNN)-based approaches to be useful for solving PS by rendering large number of images of various surfaces under numerous light and material configurations. They often parametrise the PS problem as normal regression from light intensity observations (i.e. observational map~\cite{ikehata2018cnn} ), effectively performing an inversion of the irradiance equation.  CNN-based approaches have been shown to outperform classical optimisation based methods \cite{6909677,Queau2016colored} mainly due to the ability of CNNs to learn how to deal with a great variety of realistic reflectances which lead classical optimisation methods into intractable computations and thus simplifications (e.g. assuming Lambertian reflection). In addition, CNNs can gain robustness to deviations from the irradiance equation such as global illumination effects (cast shadows, self reflections) if the training data includes them \cite{ikehata2018cnn}. This can be achieved using 3D creation suite (like Blender \cite{blender}) which can render data containing that level of realism. However, exhaustively sampling global illumination effects (which are a function of the overall surface geometry) requires a huge number of meshes to be rendered. Furthermore, the rendering requirements grow exponentially if the rendered data are to be covering all the  materials/lights configurations as well. The rendering computational cost can be reduced by rendering multi-material objects (\cite{ikehata2018cnn}) to the detriment of realistic ray-traced self reflections. Finally, it is noted that rendering full objects is computationally expensive, therefore relatively slow and somehow inefficient, as there is a large amount of correlation among neighbouring pixels (especially for shadow/self reflection patterns).

In order 
to maximise the combinations of sampled materials, lights and normal directions, instead of pre-rendering training data, we train our CNN with highly efficient, independently generated pixelwise, 
observational maps. This allows to widen the training data variation as all of these parameters (e.g materials) can be sampled independently for every data point. Moreover, we show how global light effects can be approximated in the observational map domain, and propose a strategy that includes variations in the maps that model ambient light, cast shadows, self-reflections and reflectance mixing in discontinuity boundaries. This strategy helps to reduce the synthetic to real gap making our data applicable for challenging real data \cite{diligentshi2016benchmark}.  

\noindent
\textbf{Contribution:} Our CNN based approach for solving PS problem has the following main contribution: we propose a per-pixel observation map generation strategy which can replace slow-to-obtain full image rendering while still allowing the network to learn global illumination effects. Furthermore, we also propose an improvement to the CNN-PS~\cite{ikehata2018cnn} architecture termed \usbest ~which benefits from the increase in the training data variation. 
Finally, we show that including the RGB channels  in the observation map can further boost performance.  

The rest of this work is divided as follows. Section~\ref{sec:relatedworks} discusses the relevant literature. Section~\ref{sec:Method} provides details of our proposed CNN approach. Sections~\ref{sec:experimental_setup} and~\ref{sec:experiments} 
describe the experiment setup and corresponding results.

\vspace{-0.1cm}
\section{Related Work}
\label{sec:relatedworks}
\vspace{-0.1cm}

\begin{figure*}[!t] 
\includegraphics[width=\textwidth]{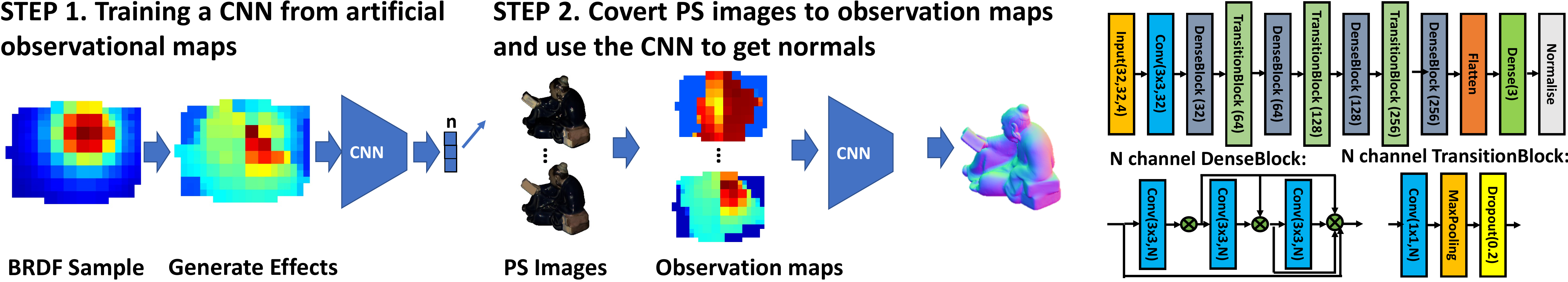}
\caption{ This figure illustrates the two key steps of our proposed approach. On the left, the network training is illustrated consisting of sampling material BRDFs and then generating synthetic observation maps (\ref{sec:data:render}) which includes modeling of global illumination and other realistic effects. In the middle, the normal estimation process is shown - for each pixel, the observation map is computed by combining the information from all PS images into a single tensor (\ref{sec:normest}). The maps are then processed by a CNN which regresses a normal (orientation) map. The complete CNN architecture is shown on the right. ReLU activation is used after each CONV layer. $\bigotimes$ denotes concatenation along the channel axis.} \vspace{-0.5cm}
\label{fig:overallmethod}
\end{figure*} 

We now provide an overview of the latest PS improvements that mostly focus on deep learning approaches. For a fairly recent survey of PS techniques, refer to \cite{AckermannG15}.

As Deep Learning (DL) has recently dominated the vast majority of ongoing research in computer vision, several approaches have been proposed for retrieving 3D geometry of objects using PS. The ability of DL-based approaches to  approximate highly non-linear mappings is highly desired in inverting non-linear and complex irradiance equations.
However, regressing very dense and accurate depth maps is not a trivial task. 
A preliminary study proposed by \cite{TangSH12a} used a Deep Belief Network \cite{Hinton09} to tackle the Lambertian PS problem. 
Recently, some approaches have been presented addressing the PS problem with DL architectures. \cite{YuS17} built layers capable of modeling photometric image formation (in a unsupervised manner) that can be embedded into existent encoder-decoder architectures for establishing correspondences among light reflections and geometry. \cite{santodeep} proposed the first network to be trained using the real-world MERL database. They simplified the generality of the PS problem by training the network with the same light direction used for testing. 

\cite{JuQZDL18} used a learning procedure for separating the RGB information from multi-spectral images when solving the RGB-PS problem. This demultiplexing procedure allowed to improve the accuracy of the reconstruction when using a minimal number of light sources. \cite{taniai2018neural} proposed an unsupervised method that does not require any training as they minimise the reconstruction loss between the rendered images and the input images at test time. This makes the approach slower with respect to usual DL based methods as the training computational time is partially transferred to the shape reconstruction pipeline. \cite{ChenHW18,chen2020deep} introduced PS-FCN, the first deep learning based approach not requiring identical set of lights at train and test time. 
Indeed, by using similar concepts, \cite{chen2019self} proposed a DL approach a more challenging scenario for the PS problem where the light sources are unknown. In this case, a two-stage modeling is used to approximate first the uniform light directions (LCNet) and then estimate the normals (NENet). \cite{zheng2019spline} proposed SplineNet to solve the Sparse PS (e.g. PS with low number of images). They employed an interpolation network to estimate the reflectance at additional light directions.
Another approach to solve Sparse PS has been presented by  \cite{wang2020non} where the lack of information due to limited amount of input images is compensated by training the network with 9x9 patches of pixels. In addition, they enforce a collocated light constrain which is derived empirically from observations in the MERL material database.  \cite{enomoto2020photometric} is able to deal with sparse configuration of lights by search and match in a  set of diverse BRDFs.

As DL has been shown to be an efficient tool for solving PS in traditional setting, that is having directional light sources and orthographic viewing geometry, recent works \cite{logothetis2020cnn, DBLPSantoWM20} exploited a CNN based approach to retrieve geometry using photometric imaging in a constrained near-field setting. Finally, \cite{ikehata2018cnn} introduced the observation map parameterisation (32 by 32 gray-scale image) that merges information of multiple lights on a single tensor allowing a fixed network to be used under a varied number of light sources. The training data was obtained by rendering 15 meshes with a dense variation of material properties under a number of light directions. The purpose of this pre-rendered training dataset was to allow the network to learn the effect of global physical phenomena synthetically generated with computer graphics. Although a large amount of data was sampled, the choice of selecting specific meshes limits the possible light-normal-material configurations and constrains the patterns of the global illumination effects (cast shadows, self reflections) which are a direct function of the global surface geometry. In addition, training on purely synthetic images without data augmentation is prone to over-fitting to the synthetic distribution with potential drop of performance in real images. 


In order to overcome these limitations, we propose a CNN-based approach for the PS problem with a better coverage of physical effects than \cite{ikehata2018cnn} without relying on pre-rendered meshes. To do so, we implemented a on observation map generation procedure sampling all the relevant parameters independently for each sample.

\section{Method}
\label{sec:Method}

This section describes the mathematical formulation of the normal estimation problem and  
provides a detailed explanation of how pixel-wise training data is generated. 
\subsection{Normal estimation}
\label{sec:normest}
Our calibrated PS approach takes as input a set of $J$ varied illumination images. The 
6
 illumination is assumed to be directional with known light directions $\vect{L}_j$ and the brightness $\phi _j$. For each pixel $p$, its value at image  $j$ is denoted as $i_{j,p}$. The objective of our method is to recover the normal $\vect{N}_p$ at each pixel $p$. This is achieved by combining all \textit{observations} of the pixel in the images with varied illumination into a single $d \times d \times 4$ map $O$ that in turn is fed into a CNN which regresses normals. A high level diagram of this procedure is shown in Figure~\ref{fig:overallmethod}. A more detailed explanation of all the steps is given below.

\noindent
\textbf{Observation map.} The concept of observational map has been introduced in \cite{ikehata2018cnn} as a way to merge information of a variable number of images into a single $d \times d$ image map. The mapping procedure follows two steps: Firstly, normalised observations $\hat{i}_{j,p}$ are computed by compensating for the light sources brightness variation, converting to grayscale (adding r,g,b components) and then dividing with the maximum (of the map): 
\begin{equation}
    \hat{i}_{j,p}=\frac{i_{j,p,\text{r}}/\phi _{j,\text{r}}+i_{j,p,\text{g}}/\phi _{j,\text{g}}+i_{j,p,\text{b}}/\phi _{j,\text{b}}}{\max _j (i_{j,p,\text{r}}/\phi _{j,\text{r}}+i_{j,p,\text{g}}/\phi _{j,\text{g}}+i_{j,p,\text{b}}/\phi _{j,\text{b}}) }.
\end{equation} 
This normalisation operation is designed to compensate for the albedo variation of different pixels, hence reducing the range of the data. Secondly, the normalised observations $\hat{i}_j$ (omitting dependence on $p$ for clarity) are placed on a square grid $O_n$ of size $d\times d $, with the location determined from the light source direction $\vect{L}_j=[l^x_j, l^y_j, l^z_j]$ as follows:
\begin{equation}
O_\text{n}\Big( \Big \lfloor d\frac{l^x_j+1}{2} \Big \rfloor,\Big \lfloor d\frac{l^y_j+1}{2} \Big \rfloor \Big)=\hat{i}_j.
\label{eq:omapn}
\end{equation}

 \begin{figure}[!t]
 \centering
\begin{tabular}{ccccccc}
  \tiny{\textbf{(a) Baseline}} \hspace{0.005\textwidth} & \tiny{\textbf{(b)Saturation}} &  \tiny{\textbf{(c)Ambient}} \hspace{0.005\textwidth} & \tiny\textbf{{(d)Shadow}} & \tiny{\textbf{(e)Reflection}} & \tiny{\textbf{(f)Discontinuity}} \\
\end{tabular} \\
\begin{sideways} {\hspace{0.1cm}\tiny Synth. $O_\text{rgb}$} \end{sideways}
\includegraphics[width=0.085\textwidth,trim={0cm 0cm 0cm  0cm},clip]{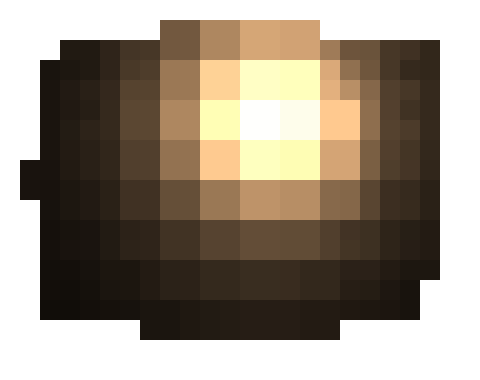} \hspace{-0.02\textwidth}
\includegraphics[width=0.085\textwidth,trim={0cm 0cm 0cm  0cm},clip]{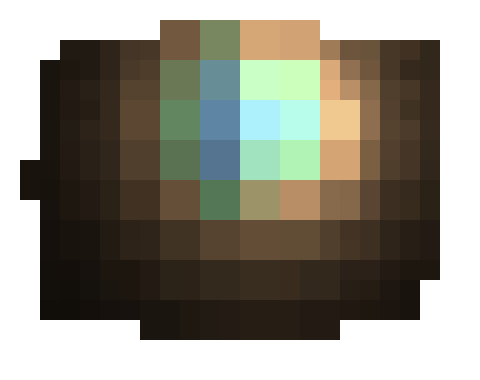} \hspace{-0.02\textwidth}
\includegraphics[width=0.085\textwidth,trim={0cm 0cm 0cm  0cm},clip]{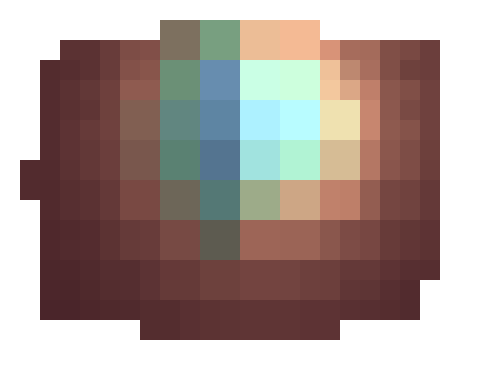} \hspace{-0.02\textwidth}
\includegraphics[width=0.085\textwidth,trim={0cm 0cm 0cm  0cm},clip]{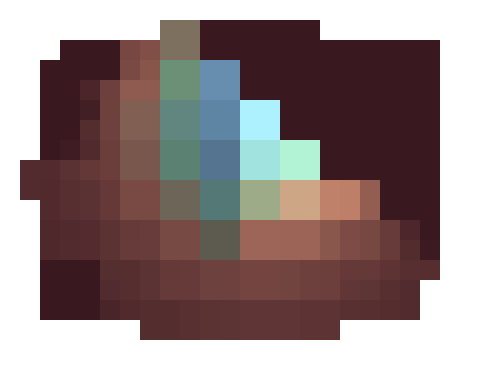} \hspace{-0.02\textwidth}
\includegraphics[width=0.085\textwidth,trim={0cm 0cm 0cm  0cm},clip]{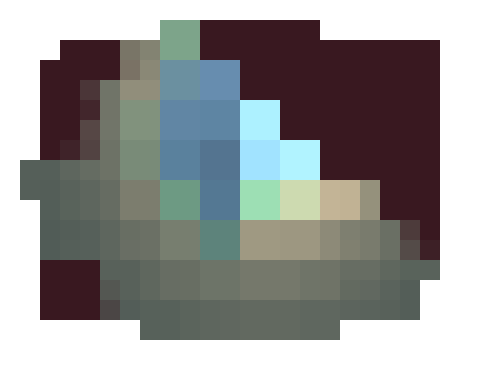} \hspace{-0.02\textwidth}
\includegraphics[width=0.085\textwidth,trim={0cm 0cm 0cm  0cm},clip]{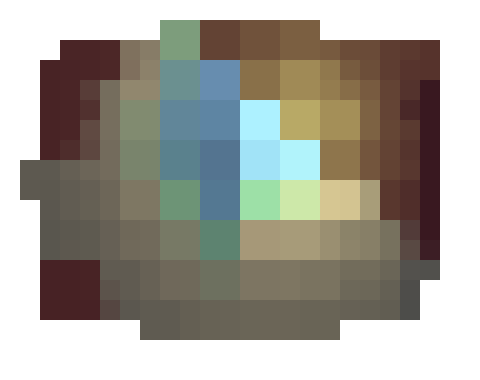}\\
\vspace{-0.2cm}
\begin{sideways} {\hspace{0.05cm} \tiny Synth. $O_n$} \end{sideways}
\includegraphics[width=0.085\textwidth,trim={0cm 0cm 0cm  0cm},clip]{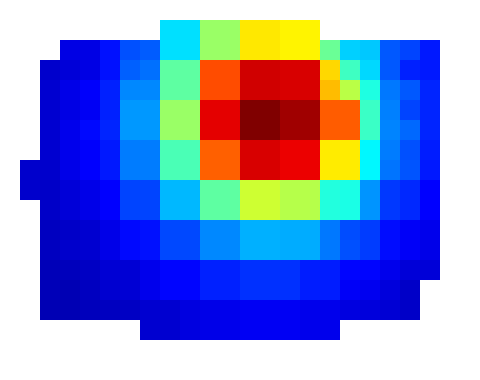} \hspace{-0.02\textwidth}
\includegraphics[width=0.085\textwidth,trim={0cm 0cm 0cm  0cm},clip]{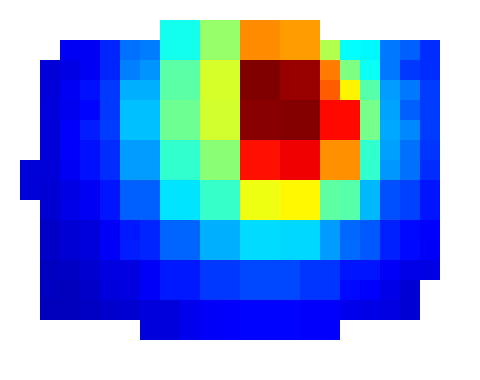} \hspace{-0.02\textwidth}
\includegraphics[width=0.085\textwidth,trim={0cm 0cm 0cm  0cm},clip]{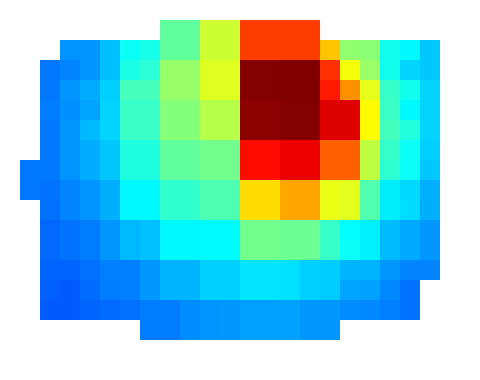} \hspace{-0.02\textwidth}
\includegraphics[width=0.085\textwidth,trim={0cm 0cm 0cm  0cm},clip]{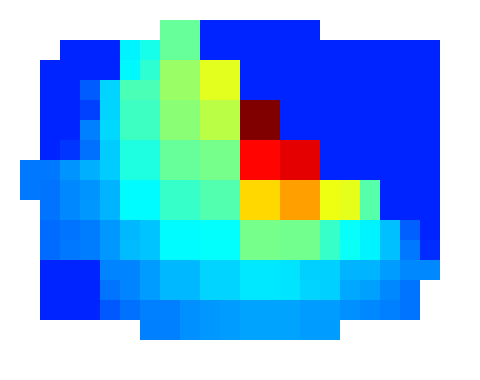} \hspace{-0.02\textwidth}
\includegraphics[width=0.085\textwidth,trim={0cm 0cm 0cm  0cm},clip]{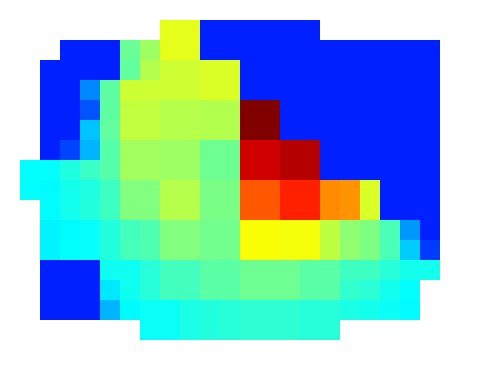} \hspace{-0.02\textwidth}
\includegraphics[width=0.085\textwidth,trim={0cm 0cm 0cm  0cm},clip]{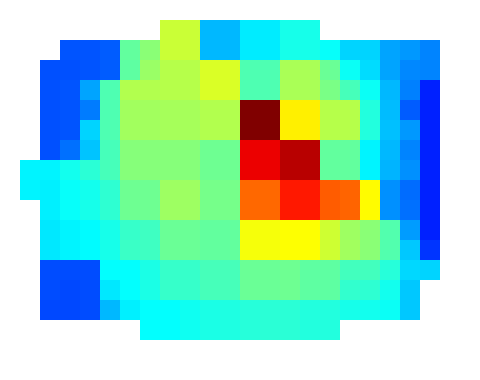}\\
\vspace{-0.2cm}
\begin{sideways} {\hspace{0.05cm} \tiny  Real $O_\text{rgb}$} \end{sideways}
\includegraphics[width=0.085\textwidth]{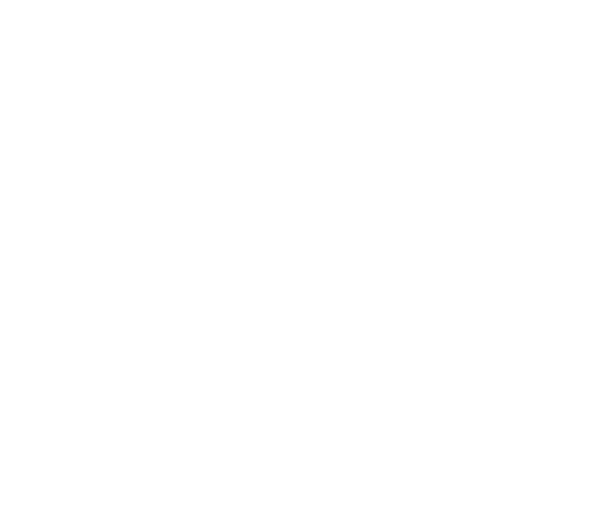} \hspace{-0.02\textwidth}
\includegraphics[width=0.085\textwidth,trim={0cm 0cm 0cm  0cm},clip]{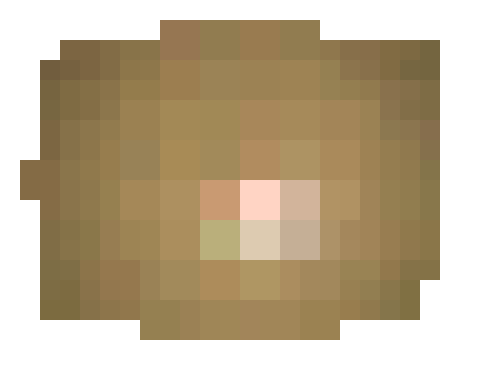} \hspace{-0.02\textwidth}
\includegraphics[width=0.085\textwidth,trim={0cm 0cm 0cm  0cm},clip]{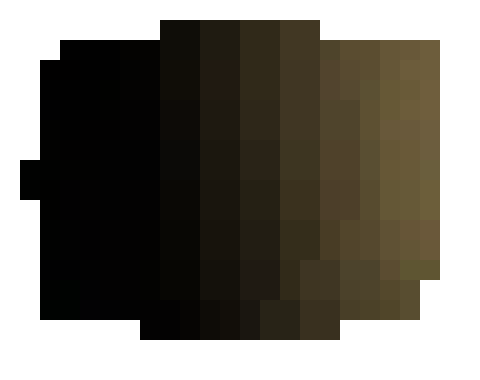} \hspace{-0.02\textwidth}
\includegraphics[width=0.085\textwidth,trim={0cm 0cm 0cm  0cm},clip]{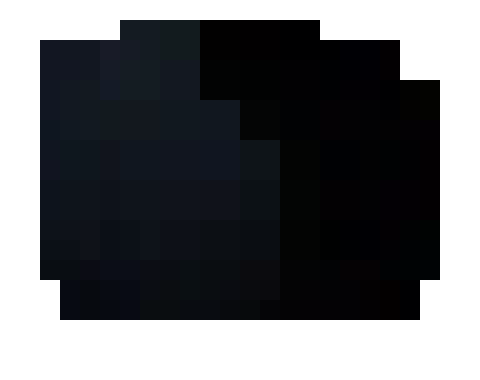} \hspace{-0.02\textwidth}
\includegraphics[width=0.085\textwidth,trim={0cm 0cm 0cm  0cm},clip]{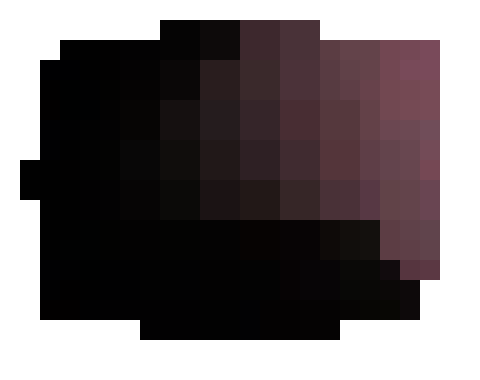} \hspace{-0.02\textwidth}
\includegraphics[width=0.085\textwidth,trim={0cm 0cm 0cm  0cm},clip]{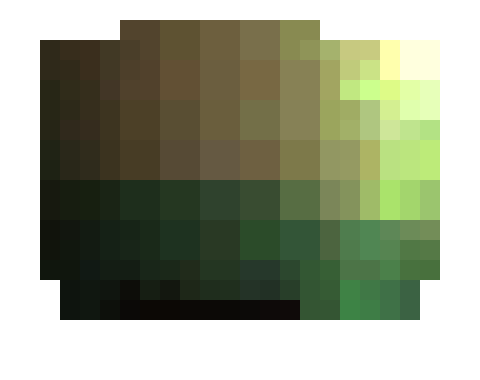}\\
\vspace{-0.2cm}
\begin{sideways} {\hspace{0.05cm}\tiny  Real $O_n$} \end{sideways}
\includegraphics[width=0.086\textwidth]{images/empty.png} \hspace{-0.02\textwidth}
\includegraphics[width=0.086\textwidth,trim={0cm 0cm 0cm  0cm},clip]{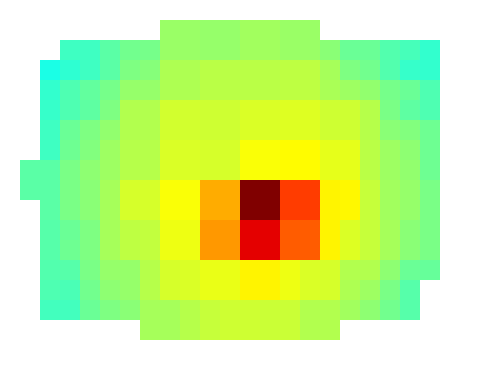} \hspace{-0.02\textwidth}
\includegraphics[width=0.086\textwidth,trim={0cm 0cm 0cm  0cm},clip]{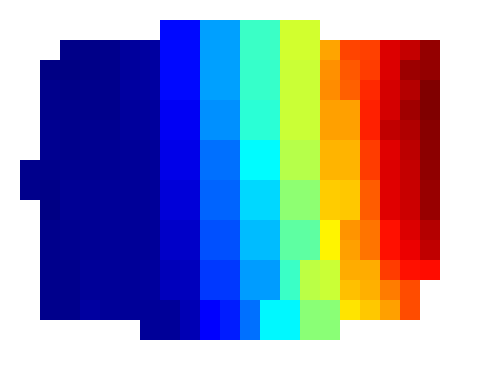} \hspace{-0.02\textwidth}
\includegraphics[width=0.086\textwidth,trim={0cm 0cm 0cm  0cm},clip]{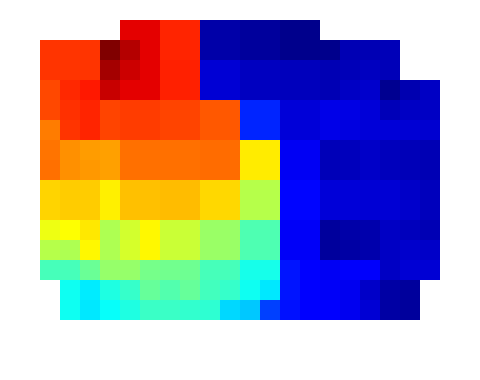} \hspace{-0.02\textwidth}
\includegraphics[width=0.086\textwidth,trim={0cm 0cm 0cm  0cm},clip]{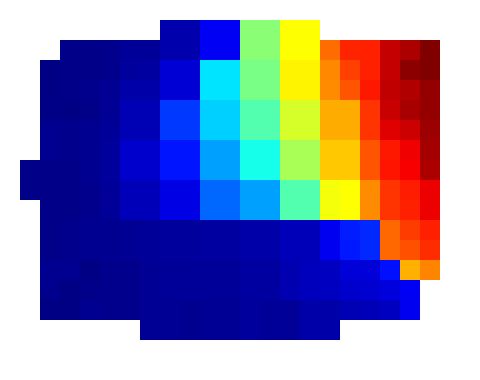} \hspace{-0.02\textwidth}
\includegraphics[width=0.086\textwidth,trim={0cm 0cm 0cm  0cm},clip]{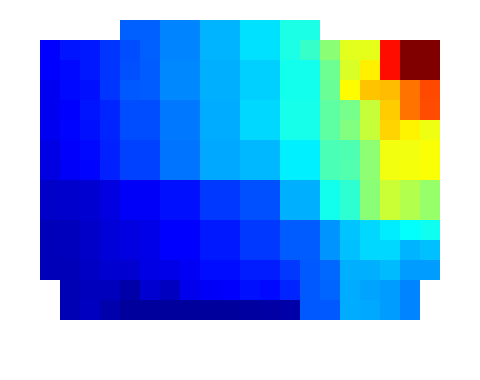}\\
\vspace{-0.2cm}
\begin{sideways} {\tiny Sample image} \end{sideways}
\includegraphics[width=0.073\textwidth]{images/empty.png}
\includegraphics[width=0.073\textwidth,trim={6.5cm 6.0cm 6cm  6.5cm},clip]{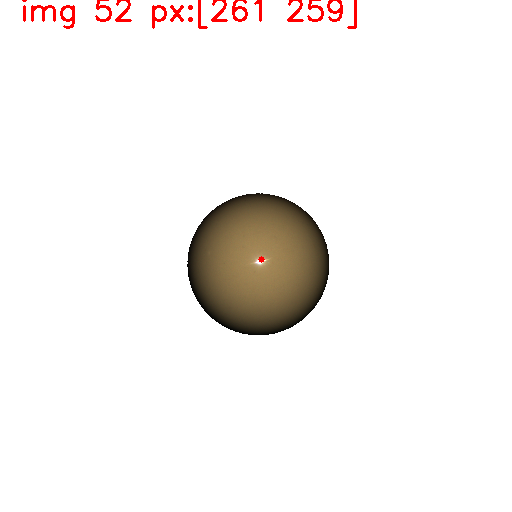}
\includegraphics[width=0.073\textwidth,trim={6.5cm 6.0cm 6cm  6.5cm},clip]{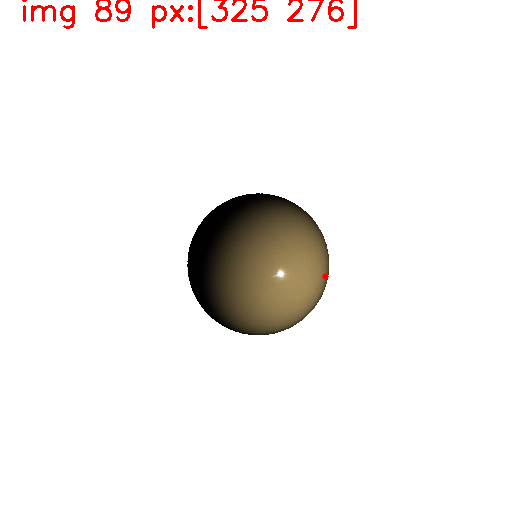}
\includegraphics[width=0.073\textwidth,trim={6.0cm 6.0cm 6.0cm 6cm},clip]{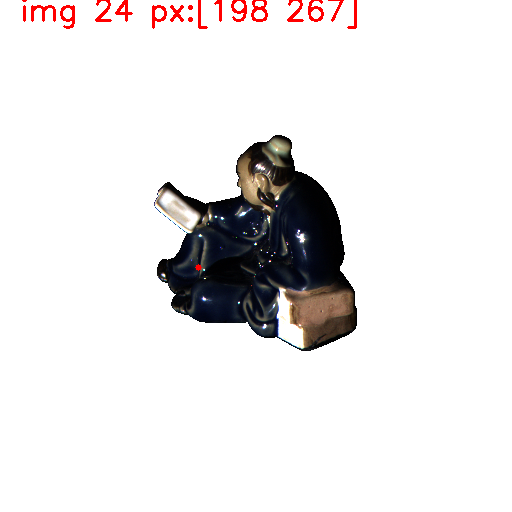}
\includegraphics[width=0.073\textwidth,trim={6.5cm 5.5cm 4cm  5cm},clip]{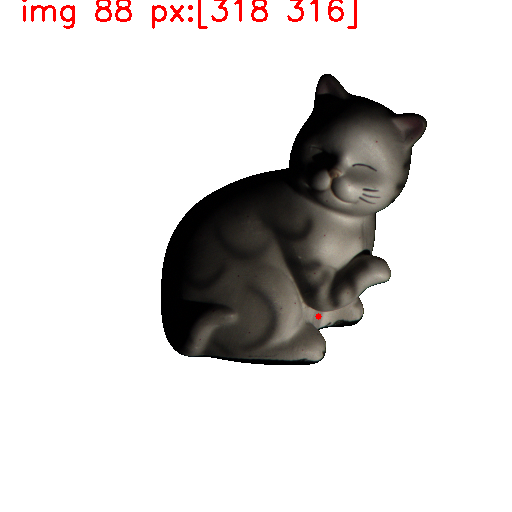}
\includegraphics[width=0.073\textwidth,trim={6.0cm 6.5cm 5.5cm  5cm},clip]{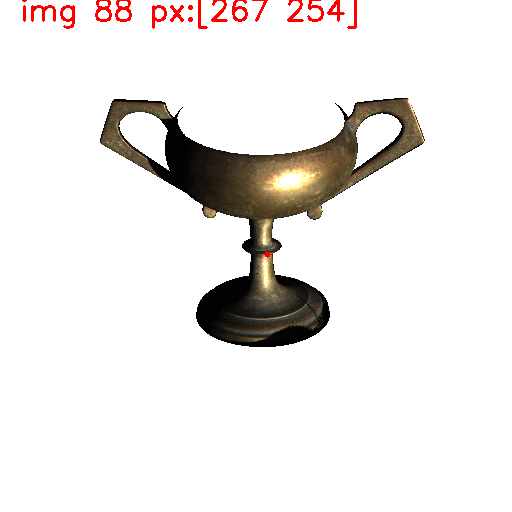}
\caption{ Demonstration of relevant effects that we model during  observation map generation. RBG maps and normalised gray (components $O_\text{rgb}$ and $O_\text{n}$ in Equation \ref{eq:omapall}) are shown. (a) is the baseline direct reflectance $r_d$ map. (b) shows the change when variable light source brightness are considered (different pixels saturate at different levels and so the normalisation with the brightness distorts the specular highlight). (c) shows the addition of ambient light that acts as an additive offset everywhere. (d) cast shadow blocks all light expect ambient/self reflection in regions of the map. (e) Self reflection can be noticed in real data by the color change: the mostly  gray cat contains red pixels at the reflection point. (f) Points at the sharp edge of the cup exhibit discontinuity (which looks like the mixing of two different maps).
   } 
  \label{fig:observationalmaps}
 \end{figure}

Note that the use of the division operation can corrupt the data in two cases. Firstly, if the maximum value is saturated, the map values are overestimated. Secondly, for very dark points, the ratio operation becomes numerically unstable and any amount of noise (or just discretisation inaccuracy) is greatly amplified. In order to overcome these limitations, we extend the observation map concept to a 3D map $O$ which also includes the RGB channels such as:
\begin{equation}
O_\text{rgb}\Big( \Big \lfloor d\frac{l^x_j+1}{2}  \Big \rfloor, \Big \lfloor d\frac{l^y_j+1}{2}  \Big \rfloor \Big)=
\begin{bmatrix}
i_{\text{r}}/\phi _{\text{r}} \\
i_{\text{g}}/\phi _{\text{g}} \\
i_{\text{b}}/\phi _{\text{b}} \\
\end{bmatrix}_j, ~~ O=[O_\text{rgb} ~; O_\text{n}]
\label{eq:omapall} 
\end{equation}
where O is a  concatenation on the 3rd axis so defining a $d \times d \times 4$ map. Finally, these observation maps are fed into a CNN which regresses surface normal $\vect{N}_p$.


\noindent
\textbf{Network training.}  As in \cite{ikehata2018cnn},  we use a CNN for regressing normals from observational maps. We use a variant of DenseNet \cite{huang2017densely}, with 16 colvolutional layers (followed by ReLU activation), 3 maxpooling and 3 dropout ones. The networks has around 4.9 million parameters and the complete diagram is shown in Figure~\ref{fig:overallmethod}. The main difference with \cite{ikehata2018cnn} is that we use 4 instead of 2 dense blocks as well as more filters per layers and we removed one of the 2 fully connected layers at the top of the network with all these changes aimed at increasing the learning capacity.

\begin{figure*}[t]
\includegraphics[width=\textwidth]{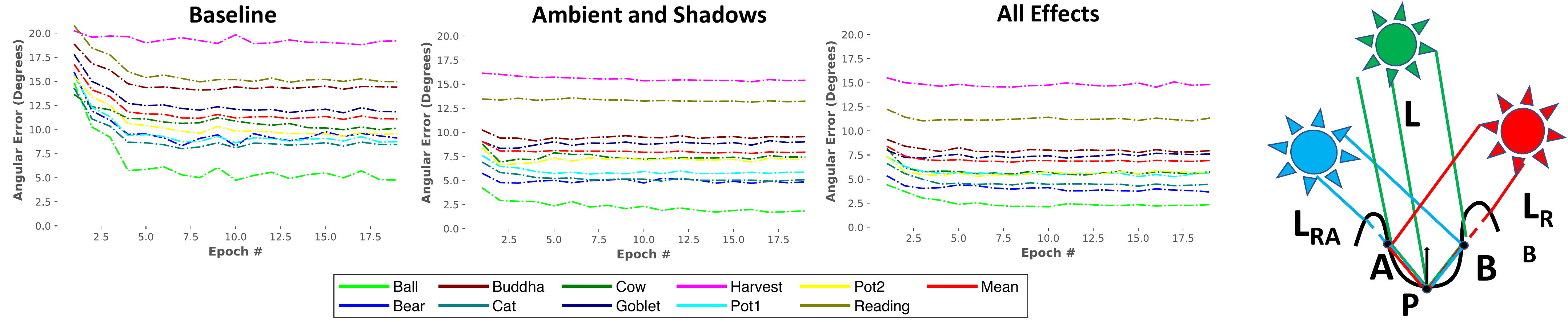}
 %
 \caption{(from the left) MAE evolution (during training) curves illustrating the  performance of the networks trained with successive effects on the DiLiGenT objects. The networks compared here are trained on maps generated with: baseline basic reflectance, ambient and global shadows only, all effects. It is observed that successive effects improve performance by sifting downwards the error curves with only notable exception being the Ball as it suffers the least from the global illumination effects. (right) Demonstration of the constraint between shadows and self reflections. In this example, red and blue light sources (assumed far away providing uniform directional illumination)  are in shadow when considering the reflection of the point P. Thus, we conclude that there are the points (A,B), located in these directions (the position along the ray does not matter in a far-field setting), which generate self reflection for the rest of the light sources. } \vspace{-0.4cm}
  \label{fig:augmentations_converge}
 \end{figure*}
 
\subsection{Data Generation}
\label{sec:data:render}

Our CNN is trained using synthetically generated  observation maps. Each  map is computed independently and  no global object rendering is performed, avoiding the need for expensive  graphics operations like surface tessellation and rasterisation. For each data point, surface normal $\vect{N}$, set of lights  $\{ \vect{L}_j \}$, albedo $\rho$  as well as material $M$ are indepenently sampled (see Section \ref{sec:experimental_setup} for parameter distributions). More specifically, $M$ is either a 9 dimensional vector containing the parameters of the  \emph{Disney} BRDF \cite{burley2012physically} (excluding anisotropy) or material index from the  MERL BRDF reflectance database \cite{Matusik2003}. Using these material parameters, a set of \emph{direct} reflectance components $\{ r_{d,j} \}=\text{B}(\vect{N},\vect{L}_j,\vect{V}_0,\rho,M)$ can be generated with either  computing the Disney \cite{burley2012physically} non-linear equation or performing table look-up for the appropriate MERL  \cite{Matusik2003} material. Note that we assume orthographic setup hence the viewing direction $\vect{V}_0=[0,0,1]$ is used for all lights $L_j$. However, real image pixel appearances deviate from pure BRDF refletance values. Such deviations include global illumination effects due to interaction of the incoming/reflected light with other parts of the surface as well as local effects such as ambient light and surface discontinuity. Therefore, in order to increase the realism of our synthetic data, a set of effects are approximated by adjusting the reflactances $r_{d,j}$ accordingly. The rest of the section explains these aforementioned effect in the context of observation map generation. Dependence on $j$ is ignored for the rest of the section for brevity.

\begin{itemize}[leftmargin=0.01cm]
\item \textbf{Cast shadows.} Cast shadows are observed in real data when a part of the surface is blocking the light, thus turning the direct reflectance to zero. This is a structured effect with very high correlation for nearby light sources. Our approximation of this effect is performed by sampling a shadow map, i.e. a binary function  $S(\vect{L})=0, \text{ if occluded } S(\vect{L})=1,$ otherwise, containing regions of occluded lights (see supplementary for details of the shadow map sampling). 
A notable difference of our work to approximating shadows with a structured dropout (e.g. \cite{li2019learning}) is that we consider the shadow map as a part of the data generation process. As it described below,  $S(\vect{L})$ is combined with other effects (self-reflections/ambient) to compute a combined pixel intensity which will be nonzero, even for shaded light sources.

\item {\textbf{Self reflections.} Self reflections occur in specular objects as a result of parts of the surface acting as auxiliary light sources. This effect can become very complicated in reality with potentially hundreds of points contributing additional light components. To estimate a computationally efficient approximation we sample up to 5 points in directions $\vect{L}_R$ and compute a single light bounce from $\vect{L}$ to $\vect{L}_R$ to $\vect{V}_0$. 
We note that in the case of directional lights far away from the surface, self reflection directions are constrained to be part of the shadow map $S(\vect{L}_R)=0$, $\forall \vect{L}_R$. This is true as any ray extending from $P$ outwards, either intersects the surface at another point $A$ or extends to infinity unintersected. In the first case, $A$ is a potential self reflection point and in the second case light can be received from a far away light source (see Figure~\ref{fig:augmentations_converge}). 
For each of these self reflection points, surface normals $\vect{N}_R$ and albedo $\rho_R$
are independently sampled but the material is assumed to be the same (see supplementary material for further justification). Then, the self reflection component for light $\vect{L}$ at reflection points $\{ \vect{L}_R \}$ is:  
\begin{equation}
 \small {r_r(\vect{L},\{ \vect{L}_R \})= \sum _{\vect{L}_R,~ \vect{L}_R \neq \vect{L} }\text{\small B}(\vect{N}_R,\vect{L},\vect{L}_R,\cdots) \text{B}(\vect{N},\vect{L}_R,\vect{V}_0,\cdots)}. 
\label{eq:selfref}
\end{equation}
Note that in the first BRDF term, the effective view vector is now $\vect{L}_R$ which is also the effective light vector in the second term (as light travels from $\vect{L}$ to $\vect{L}_R$ to $\vect{V}_0$). We note that this single bounce ray-tracing  does not fully compensate for the case of multiple light bounces, subsurface scattering or other more complicated global light transform paths. Finally, we note that self reflections alters the color of the pixel (the multiplication of the BRDFs in Equation~\ref{eq:selfref} will increase the color saturation) and thus the inclusion of the RGB channels is further motivated in order to give information to the network to disambiguate the self reflection component from the main one. }  

\item{\textbf{Surface discontinuity.} It is common to assume that each pixel corresponds to the reflection of a single surface point with a specific normal (e.g. in differential approaches like \cite{MeccaQLC2016} which assume continous surfaces). However, in practice, pixels have a finite size and thus it is likely that they record the reflectance of multiple surface points with potentially different surface normals. This effect is mostly relevant at surface discontinuity points, such as occlusion boundaries and sharp edges. 
As the BRDF is a nonlinear function of $\vect{N}$, this mixing effect needs to be accounted for as well. Our implementation  approximates this effect by sampling $t\in \{1,2,3\}$ normals $\vect{N}_k$ per pixel (85\% of pixels get $t=1$ to not have this effect) and then average out the respective reflectances (both direct $r_d$, which may be blocked by shadow, and due to self reflection $r_r$) to compute an overall reflectance $r_T=\sum _{k} \frac{r_d(\vect{N}_k)S(\vect{L}) + r_r(\vect{N}_k) }{t}$ }.

\item { \textbf{Ambient light.} Most real images contain some amount of ambient light mostly due to light dispersing into the atmosphere and reflecting on other objects in the environment. Even if the PS images are captured in a dark room with no reflective objects, this effect still persists even though it can be very small ($\approx 0.1\%$ of the maximum intensity). This effect is usually modeled (e.g. \cite{logothetis2016near}) as a constant reflection $a$. We note that this constant reflection has high correlation with  surface albedo and it is also diminished at very oblique angles (along with  most of the reflection) so we sample such as $a \propto \rho \vect{N} \cdot \vect{V}_0$. We finally, include an additional small (up to $1e-4$) additive uniform noise $n_{AU}$ component to account for any additional light arriving to the camera (e.g. reflecting of atmospheric dust).}
\begin{figure}[t]
\includegraphics[height=0.125\textwidth,trim={12.5cm 1.0cm 1.0cm  0.5cm},clip]{./images/colorbar_up.pdf} \hspace{-0.3cm}
\includegraphics[height=0.125\textwidth,trim={7.5cm 5.5cm 7.5cm  0.5cm},clip]{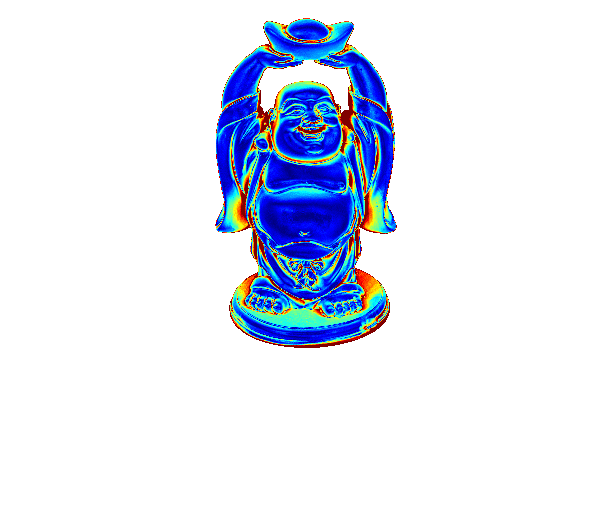} 
\includegraphics[height=0.125\textwidth,trim={7.5cm 5.5cm 7.5cm  0.5cm},clip]{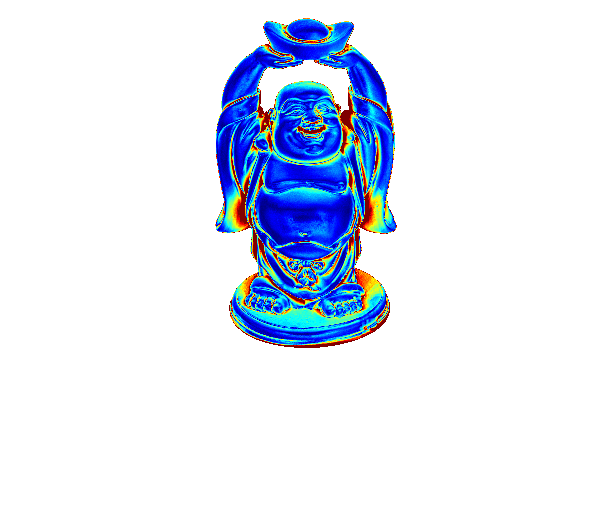} 
\includegraphics[height=0.125\textwidth,trim={7.5cm 5.5cm 7.5cm  0.5cm},clip]{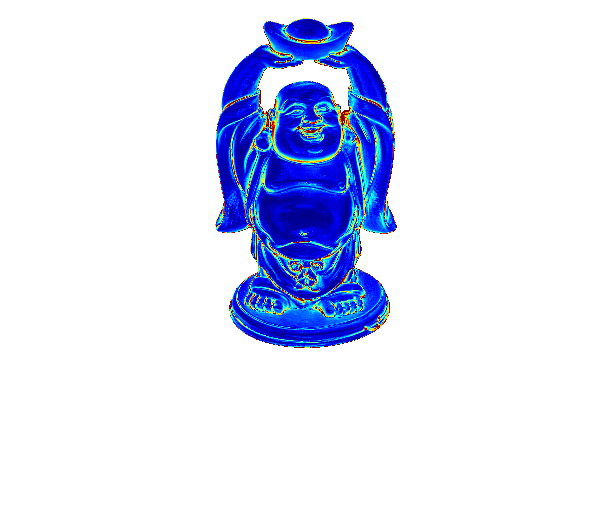}
\includegraphics[height=0.125\textwidth,trim={7.5cm 5.5cm 7.5cm  0.5cm},clip]{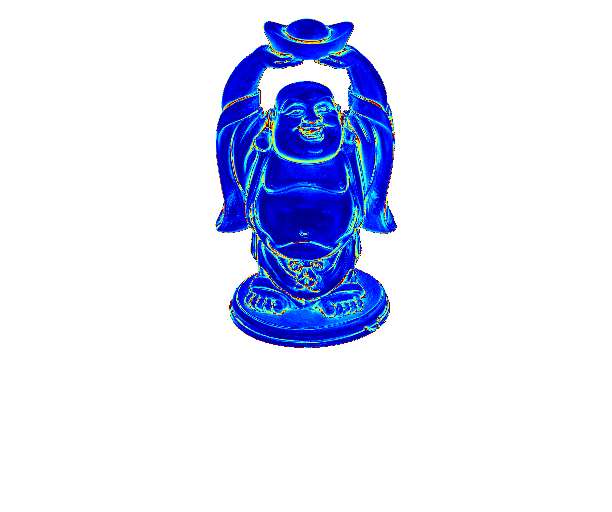}
\includegraphics[height=0.125\textwidth,trim={7.5cm 5.5cm 7.5cm  0.5cm},clip]{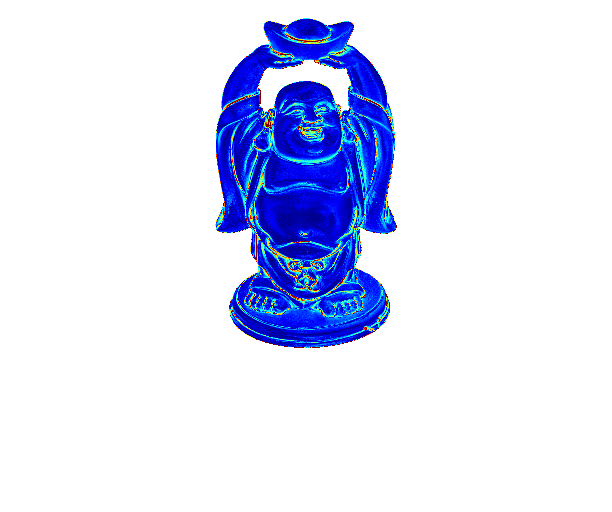}
\includegraphics[height=0.125\textwidth,trim={7.5cm 5.5cm 7.5cm  0.5cm},clip]{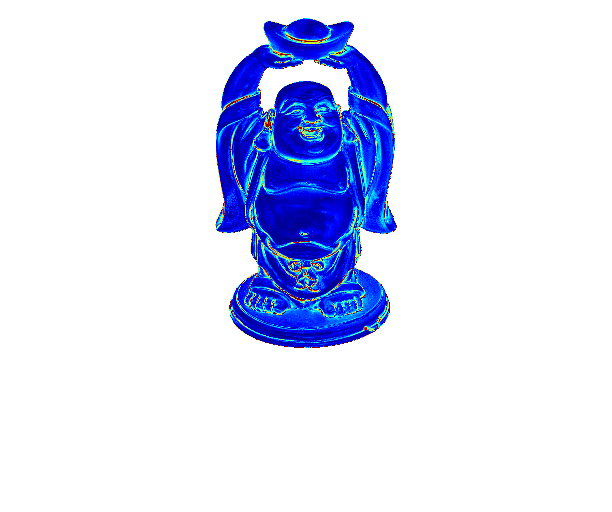}
\begin{tabular}{cccccccc}
\StackC[\tiny]{~~\\ ~~} &
\StackC[\tiny]{(a)Baseline \\ $14.61^o$} &
\StackC[\tiny]{(b)Saturation \\ $14.45 ^o$} &
\StackC[\tiny]{(c)Shadow  \\ $9.33^o$} & 
\StackC[\tiny]{(d)Ambient \\ $8.38^o$} &
\StackC[\tiny]{(e)Reflection \\ $7.97^o$} &
\StackC[\tiny]{(f)Discontinuity \\ $7.78^o$} 
\end{tabular}
 \caption{Demonstration of the impact of incremental modeling of effects  in the performance of our \usik ~on the `Buddha' of DiLiGenT real dataset. (a) Shows the result with the baseline network. For the rest of the effects, most of the improvement is at: (b) specular highlights middle of pot, (c) top of head, (d) significantly in most concave regions, (e) middle of head (f) sleeves.}
 \label{fig:incremental_increase}
\end{figure} 
\item{\textbf{Camera Noise.} Real cameras are prone to noise which can reasonably approximated as Gaussian. We include two components: a multiplicative $n_{MG}$ and an additive one $n_{AG}$. In practice, these components are small so we assume standard deviations of $1e-4$ (means are 1 and 0 respectively). }
\item{\textbf{Miscellaneous.} Finally, we include a multiplicative uniform noise component $n_{MU}$ aimed to address miscellaneous unmodeled physical effects. These include light source brightness calibration \cite{logothetis2017semi} uncertainty and near light attenuation (as in reality point light sources are not infinitely far away) which affect pixel brightness in a multiplicative way. We empirically observed that this was the most important noise component with an optimum value of 5\%. }
\item{\textbf{Saturation Variation.} Different real light sources have varied brightness $\phi _J$. The observation map parameterisation aims to compensate for this variation through dividing by $\phi _j$. However, in practice,  pixel saturation makes this compensation imperfect and thus needs to be augmented for. The practical implementation involves sampling a brightness value $\phi _j$, multiply the reflectance with this value, apply the rest of the augmentation and then apply discretisation and saturation i.e. $D(x)=(\text{uint}_{16}(2^{16}x))/2^{16}$.  
Thus for saturated pixels, the final division to create an observation map does not fully compensate the light source brightness\footnote{In Diligent, $\phi$ varies between $0.28$ and $3.2$ hence saturated values map to $0.31-3.57$ which is a fairly big variation.}. Note that as the brightness is different for different channels, this results into specular highlights not being completely white in the brightness compensated images.}
\end{itemize}
Combining all the above effects the overall generated pixel intensity $i$ is calculated as follows:
\begin{equation}
i_j= D \Big(  (r_{T,j} + a) \phi _j n_{MU,j} n_{MG,j} + n_{AU,j}+ n_{AG,j}  \Big ).
\label{eq:augm}    
\end{equation}

\noindent
Finally, $\{ i_j \}$ are converted into an observation map as explained in Section \ref{sec:normest}. Visual illustration of these effects in real image maps and our synthetically generated are shown in Figure~\ref{fig:observationalmaps}. Note that the synthetic maps at Figure~\ref{fig:observationalmaps} are generated with Diligent \cite{diligentshi2016benchmark} lights to be comparable to real ones - we used random lights during train time. Detailed explanation for all the relevant hyperparameters can be found in the supplementary material.  

\begin{table}[t]
\setlength{\tabcolsep}{0.75pt} 
\begin{center}
\resizebox{1.0\columnwidth}{!}{%
\begin{tabular}{ | l | c  c  c  c  c  c  c  c  c  c  | c | }
\hline
Effects & Ball & Bear & Buddha & Cat & Cow & Goblet & Harvest & Pot1 & Pot2 & Reading & AVG  \\ \hline
Baseline         & 5.8 &  9.8 & 14.6 &  8.8 & 10.1 & 12.5 & 18.9 &  9.4 &  9.7 & 15.4 & 11.48 \\ \hline
$+$Sat        & 3.8 &  8.3 & 14.5 &  7.9 & 10.1 & 12.1 & 19.4 &  8.9 &  9.6 & 15.1 & 10.97 \\ \hline
$+$Shadow        & 1.7 &  4.8 &  9.3 &  4.9 &  7.1 &  8.8 & 15.3 &  5.8 &  7.0 & 13.1 &  7.78 \\ \hline
$+$Ambient       & 2.1 &  4.2 &  8.4 &  4.6 &  5.8 &  7.8 & 14.9 &  5.6 &  6.0 & 12.5 &  7.18 \\ \hline
$+$Reflection    & 2.4 &  3.7 &  8.0 &  4.5 &  5.8 &  7.6 & 14.5 &  5.4 &  5.6 & 10.9 &  6.83 \\ \hline
$+$Discontinuity & 2.2 &  3.7 &  7.8 &  4.3 &  5.5 &  7.6 & 15.0 &  5.4 &  5.7 & 10.9 &  6.79 \\ \hline
\usbest~$O_n$ & 1.9 &  3.7 &  7.6 &  4.5 &  5.4 &  7.0 & 13.3 &  5.2 &  5.2 & 10.5 &  6.43 \\ 
\usbest $+O_{rgb}$ & 2.0 & 3.6 & 7.6 &4.4 & 4.7 & 6.9 & 13.1 & 5.1 & 5.1 & 10.3 & 6.28 \\ \hline
\end{tabular}}
\vspace{0.05cm}
\caption{ Ablation study of the components of the several modelled physical effects in the accuracy of our \usik ~ on the real data of DiLiGenT  \cite{diligentshi2016benchmark}. It is observed that performance is almost monotonic for all objects. One notable exception is the Ball which has no reflections and ambient so including these effects decrease the performance for this object. Last 2 lines are obtained using the improved architecture \usbest ~ which further reduces the normal error (both with normalised maps as well as with the inclusion of the RGB channels). \label{table:aug}}\vspace{-0.4cm}
\end{center}
\end{table}

\begin{figure}[t]
\noindent
\resizebox{1.0\columnwidth}{!}{%
\begin{tabular}{ | l | c c c c c c c c c c c c| l | }
 \hline
Method & \multicolumn{2}{c}{Sphere 17} & \multicolumn{2}{c}{Turtle 17} & \multicolumn{2}{c}{Paperbowl 17} &
 \multicolumn{2}{c}{Sphere 305} & \multicolumn{2}{c}{Turtle 305} & \multicolumn{2}{c|}{Paperbowl 305} & AVG  \\ 
~ & (S) & (M) & (S) & (M) & (S) & (M) & (S) & (M) & (S) & (M) & (S) & (M) & ~ \\
 \hline
Baseline \cite{Woodham1980} & 4.6 & 35.4 & 12.3 & 39.7 & 29.4 & 39.3 & 5.0 & 44.5 & 12.6 & 40.2 & 28.8 & 37.0 & 27.4  \\ 
\hline 
 CNN-PS\cite{ikehata2018cnn}, K=1 & 4.9 & 11.9 & 12.9 & 20.2 & 22.2 & 36.8 & 1.6 & 2.0 & 4.2 & 6.5 & 6.8 & 11.1 & 11.8   \\ 
\hline
\usbest ,~K=1 & 1.78 &  4.52 & 7.45 & 13.68 & 22.14 & 36.01 & 0.68 & 6.45 & 3.46 & 10.66 & 7.32 & 22.19 & 11.36 \\
\hline
\end{tabular}
} 
\vspace{0.1cm} \\
\begin{minipage}{0.98\columnwidth}
\begin{sideways} {\tiny Sample PS image} \end{sideways} 
\includegraphics[height=0.07\textheight,trim={3.5cm 0cm 3.15cm  0cm},clip]{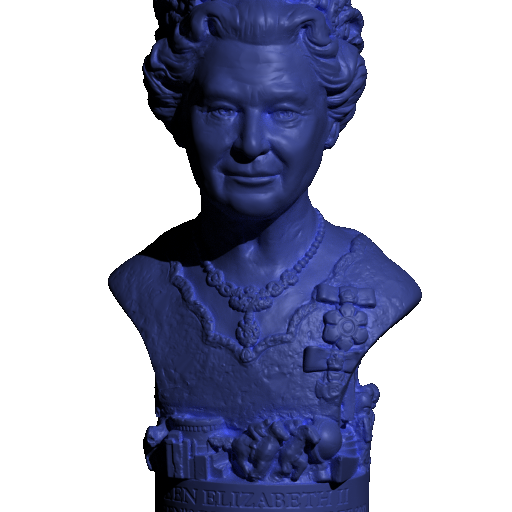} 
\includegraphics[height=0.07\textheight,trim={0.5cm 2.5cm 0.0cm 2.0cm} ,clip]{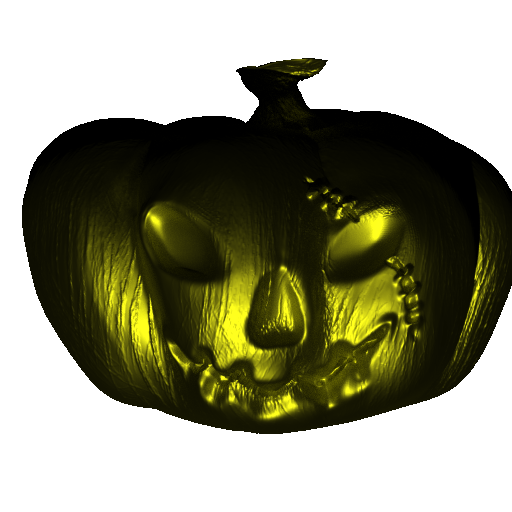}
\includegraphics[height=0.07\textheight,trim={1cm 5cm 1cm  5cm},clip]{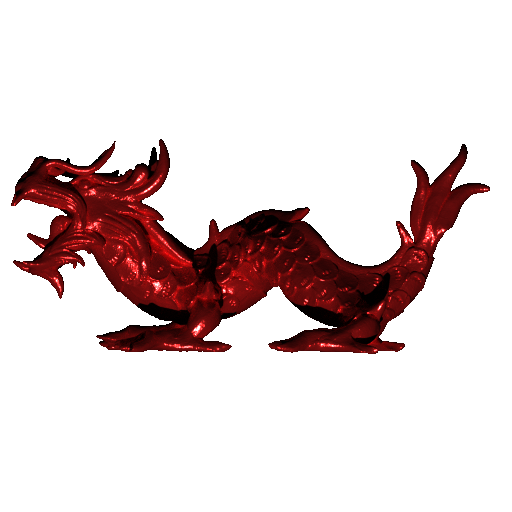} 
\includegraphics[height=0.07\textheight,trim={1cm 0cm 1cm  0cm},clip]{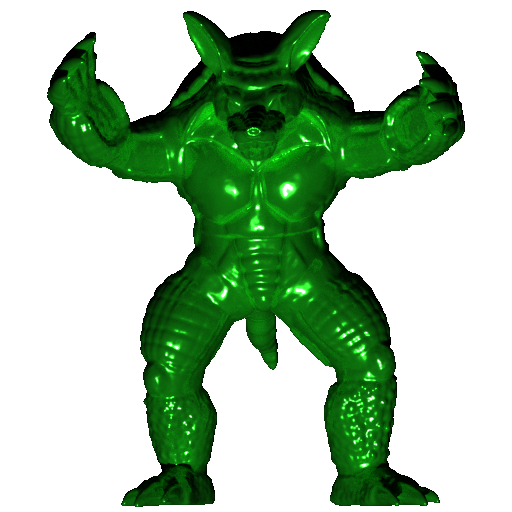}\\
 \StackC[\scriptsize]{\hspace{0.25cm} QUEEN \\ \hspace{0.3cm}(Specular)}  \hspace{0.4cm}
\StackC[\scriptsize]{PUMPKIN \\ (Metallic)} \hspace{0.9cm}
\StackC[\scriptsize]{DRAGON \\ (Intermediate)} \hspace{0.9cm}  
\StackC[\scriptsize]{ARMADILLO \\ (Intermediate)} \\

\begin{sideways} {\tiny CNN-PS\cite{ikehata2018cnn}} \end{sideways} 
\StackC{\includegraphics[height=0.07\textheight,trim={3.5cm 0cm 3.15cm  0cm},clip]{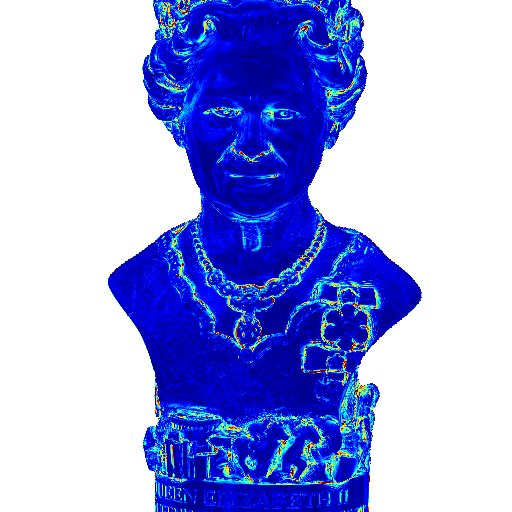} \\ \scriptsize $5.69^o$}
\StackC{\includegraphics[height=0.07\textheight,trim={0.5cm 2.5cm 0.0cm  2.0cm},clip]{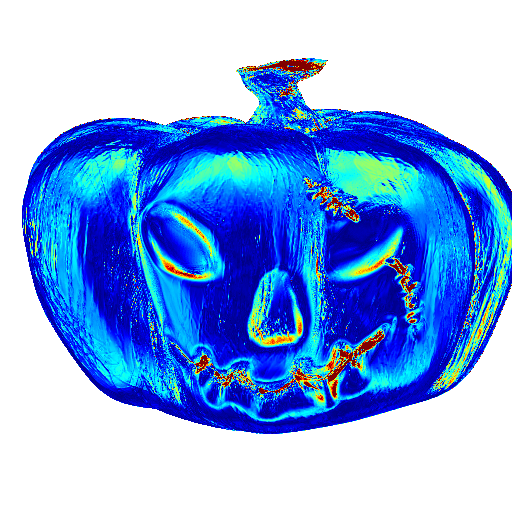} \\ \scriptsize $9.89^o$}
\StackC{\includegraphics[height=0.07\textheight,trim={1cm 5cm 1cm  5cm},clip]{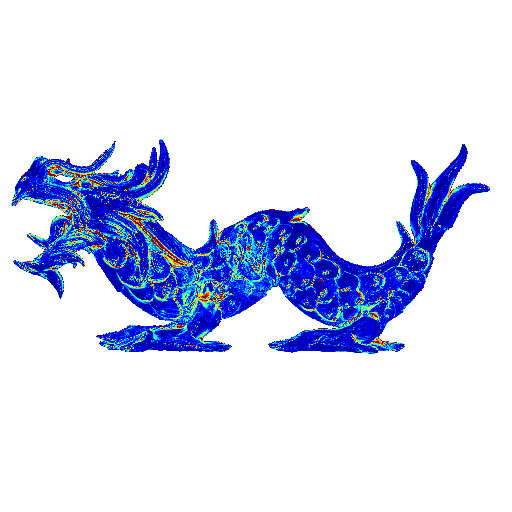} \\ \scriptsize $9.93^o$}
\StackC{\includegraphics[height=0.07\textheight,trim={1cm 0cm 1cm  0cm},clip]{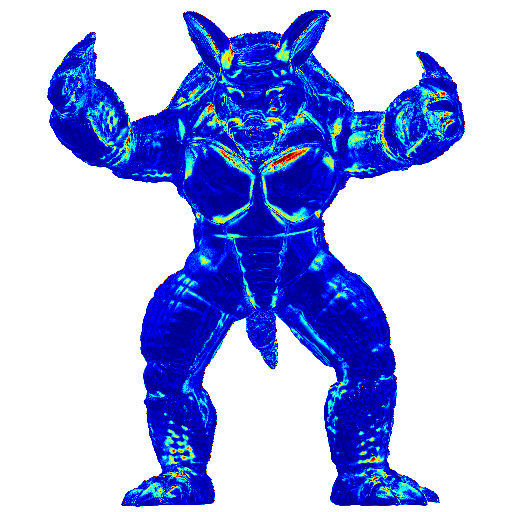}\\ \scriptsize $6.09^o$}

\begin{sideways} {\tiny \usbest} \end{sideways} 
\StackC{\includegraphics[height=0.07\textheight, trim={3.5cm 0cm 3.15cm  0cm},clip]{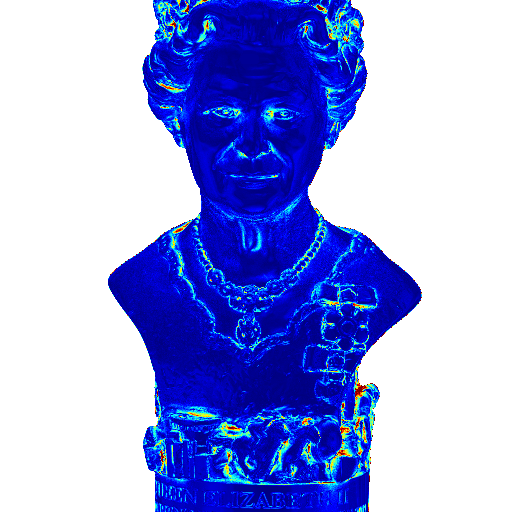}\\ \scriptsize $3.69^o$}
\StackC{\includegraphics[height=0.07\textheight,trim={0.5cm 2.5cm 0.0cm  2.0cm},clip]{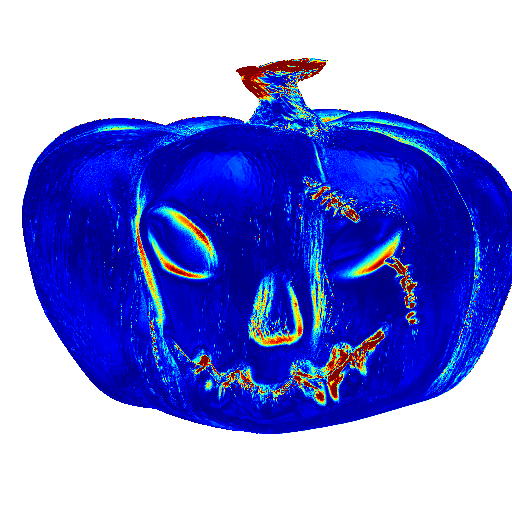}\\ \scriptsize $5.33^o$}
\StackC{\includegraphics[height=0.07\textheight,trim={1cm 5cm 1cm  5cm},clip]{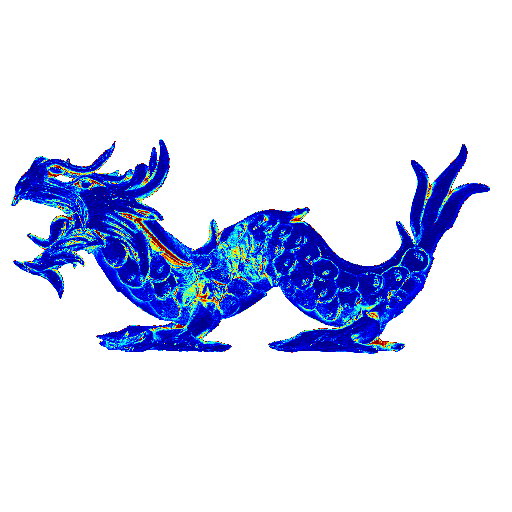}\\ \scriptsize $7.18^o$}
\StackC{\includegraphics[height=0.07\textheight,trim={1cm 0cm 1cm  0cm},clip]{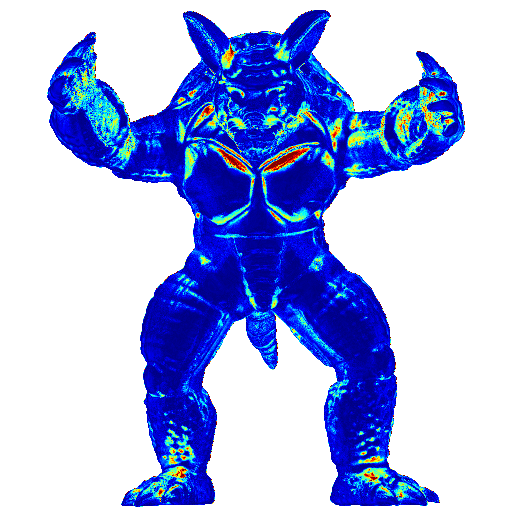}\\ \scriptsize $4.40^o$} \\

\end{minipage}\begin{minipage}{0.0425\textwidth}
 \hspace{-0.2cm}
 \includegraphics[width=0.95\textwidth,trim={12.5cm 1.0cm 1.0cm  0.5cm},clip]{./images/colorbar_up.pdf}
\end{minipage}
\\
\vspace{-0.2cm}

\caption{Comparison of our \usbest ~  with CNN-PS~\cite{ikehata2018cnn} on synthetic, globally rendered objects. The table at the top shows performance of Cycles-PS (from \cite{ikehata2018cnn}) and on the bottom on our, uniform material objects. The proposed approach outperforms the competition on both datasets, especially on the uniform material objects (average MAE $7.90^o$ vs $5.15^o$).   }
 \label{fig:blen} 
\end{figure}


\section{Experimental Setup}
\label{sec:experimental_setup}
This section describes our experimental setup including the datasets used, training and evaluation procedure.
\noindent
\textbf{Datasets.} We use three synthetic and one real dataset for evaluation. The real dataset used for the experiments is  DiLiGenT \cite{diligentshi2016benchmark} consisting of 10 objects of varied materials and geometry. For each object, 96 images ($612 \times 512$ px) are provided along with ground truth light source directions, brightness and normal maps. For the `Bear' object, the first 20 images are corrupted and thus are removed as reported by \cite{ikehata2018cnn}. We perform full lights as well as sparse lights evaluation, the later consisting of 10 random subsets of 10 lights. In addition,  we consider two synthetic datasets rendered with Blender (using the Cycles render engine) which performs realistic computation of global illumination effects. The first dataset is  Cycles-PS-Test \cite{ikehata2018cnn} containing 3 objects. Each object is rendered in a multi-material setting  (each superpixel has a different random material) from either a specular or metallic distribution. As this material distribution is unrealistic, we generated a second dataset where we rendered four single material/albedo objects namely QUEEN, PUMPKIN, ARMADILLO, DRAGON (see Figure~\ref{fig:blen}). These objects are non-convex and were rendered with Blender (16 bit 512x512 px images) including realistic global illumination effects using the 96 light sources from DiLiGenT. 
Finally, the ability of the network to learn materials is evaluated on a synthetic dataset of spheres using the MERL materials \cite{Matusik2003}. For all 100 materials, we (pixelwise) rendered 96 spheres using the DiLiGenT lights.
\begin{figure}[t]
\hspace{-0.2cm}
 \includegraphics[height=0.2\textwidth,trim={0.0cm 0.5cm 0.5cm  0.2cm},clip]{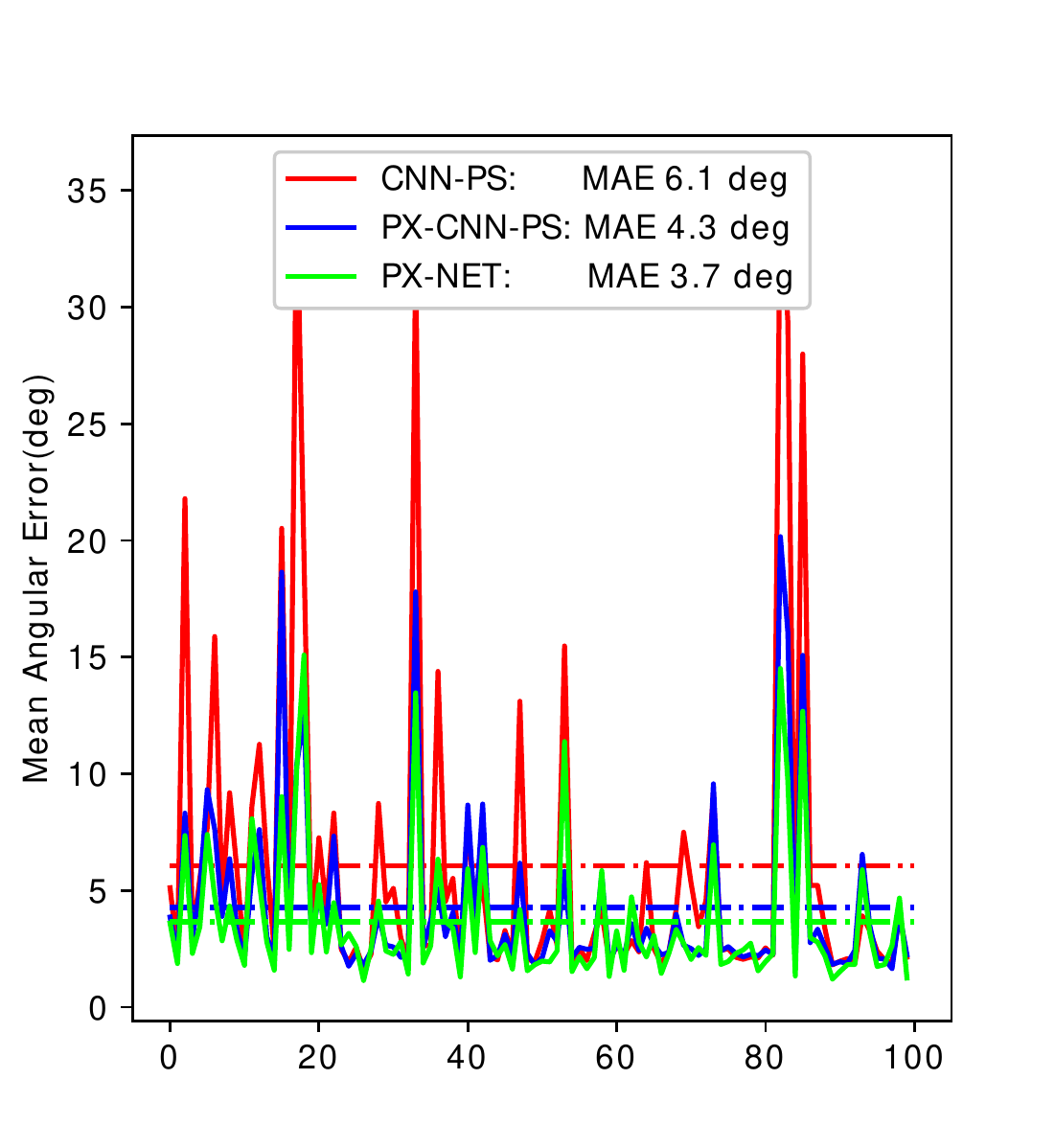}
  \includegraphics[height=0.2\textwidth,trim={0.5cm 0.5cm 0.5cm  0.2cm},clip]{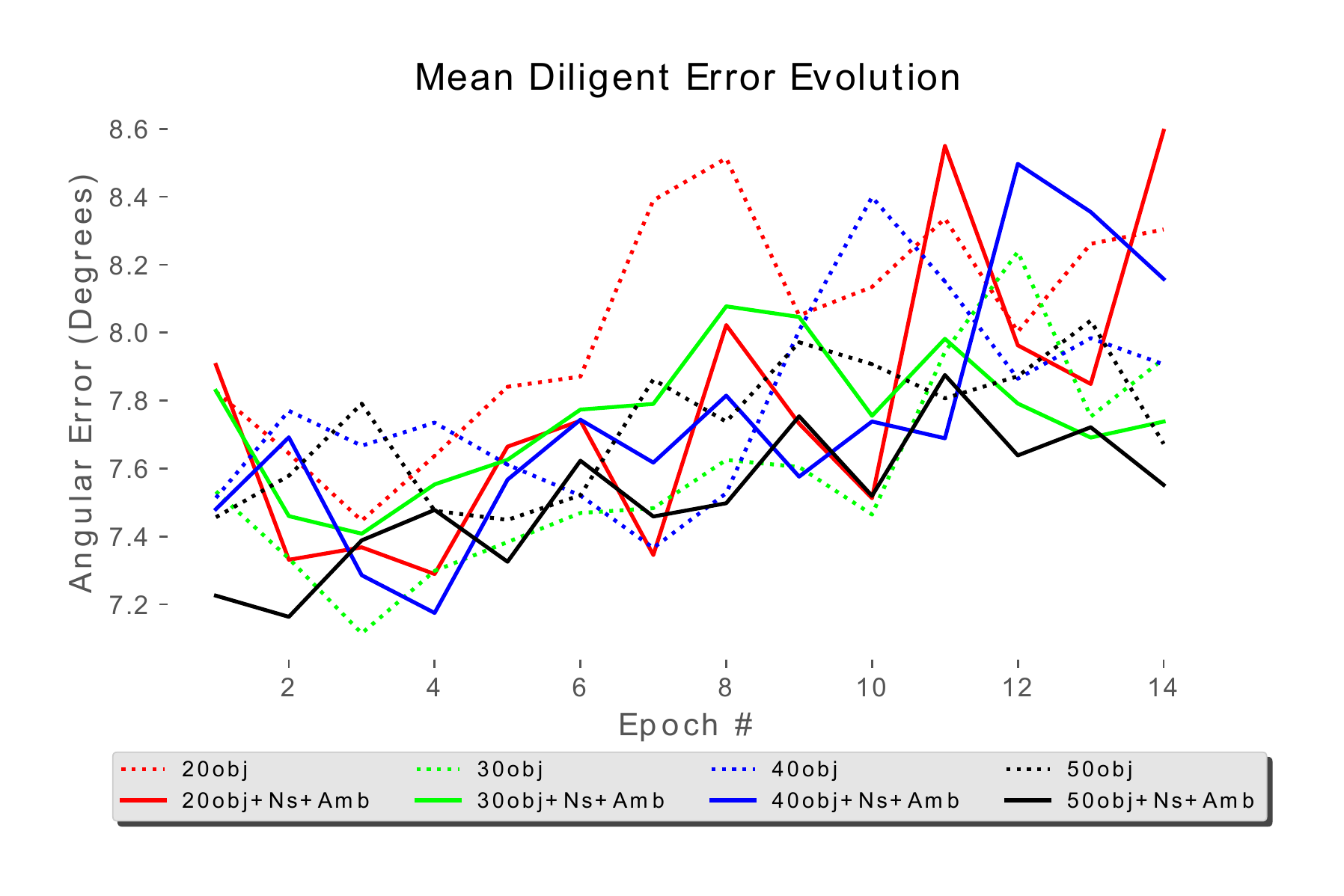}
 \vspace{-0.5cm}
  \caption{ (left) Comparison of CNN-PS \cite{ikehata2018cnn}  with our two networks  (\usik ~,\usbest ) tested on synthetic MERL images which were rendered with the DiLiGenT lights. All results are shown for a K=1 prediction with corresponding mean errors $6.1,~ 4.3, ~ 3.7$ illustrated as horizontal lines. (right) Test-time accuracy (on DiLiGenT~\cite{diligentshi2016benchmark} dataset) evolution of CNN-PS~\cite{ikehata2018cnn} network when trained on an increased number of objects (20, 30, 40, 50) using original data generation and training protocol using (dashed lines). Solid lines indicate the performance of the corresponding networks trained with additional ambient light, camera noise and multiplicative noise effects.}  
 \label{fig:mrl}
\end{figure}

\noindent
\textbf{Training details.} Baseline experiments are performed using the exact architecture of \cite{ikehata2018cnn}. This version will be termed \usik. The only difference compared to \cite{ikehata2018cnn} for this version is the training data that was made using our data generation procedure explained in Section~\ref{sec:data:render}. The purpose of \usik ~is to show that our data generation strategy can compensate for the real and global effects and even achieve state of the art results on the dense light setting (\usik ~with K=10 is only outclassed by our \usbest , see Table~\ref{tab:main}). Final experiments are performed with our modified version of the architecture called \usbest ~which achieves significantly better results.Note that a single \usbest ~ was trained for all dense experiments and another one for sparse ones.

\noindent
\textbf{Implementation.} The network was implemented in Keras of Tensorflow 2.0. The data generation engine was implemented in Python and C++ with the only external dependence being OpenCV for basic vector algebra and i/o. We trained the network using the mean angular error (MAE) loss function, which is also the evaluation metric for all experiments. For the predicted normals $\vect{n}_p$ and ground truth ones $\vect{n}_t$, MAE is  computed as: $|\text{atan2}(||\vect{n}_t \times \vect{n}_p||, \vect{n}_t \cdot \vect{n}_p)|.$

\noindent
\textbf{Hyper parameters.} The training batch size was set at 2400 with 5000 batches per epoch (12M maps). We trained for 20 epochs (which is enough for convergence Figure~\ref{fig:augmentations_converge}) using the default settings of the Adam \cite{kingma2014adam} optimiser. The train time was around 7 hours for \usik ~and 15 hours for \usbest ~on a NVIDIA GeForce RTX 2080Ti.  The  light distribution was set to 50-1000 random lights (sampled uniformly with elevation angle ranging from $0^o$ to $70^o$) in order to have a fair comparison with~\cite{ikehata2018cnn}. The sparse light setup uses 10 random lights up to $45^o$ (to match \cite{li2019learning}). The exact hyperparameters of the data generation procedure are described in the supplementary material. 


\noindent
\textbf{Rotation psedo-invariance:} \cite{ikehata2018cnn} notes that observation maps can be rotated in order to perform a test time augmentation (using 10 rotation which is termed as K=10).  If this augmentation is not used (which is the default choice in the paper unless otherwise stated), the single network evaluation is termed K=1.

\section{Experiments}
\label{sec:experiments}
In this section we present 
experiments showing state of the art performance in the datasets described in Section~\ref{sec:experimental_setup}.

\noindent
\textbf{Ablation of realistic effects modeling.} Our first experiment aimed at evaluating the effect of the incremental modeling to demonstrate how the network trained with per-pixel data can outperform the network trained with globally rendered training data. For that, we first trained a series of networks with the exact same architecture of CNN-PS \cite{ikehata2018cnn}, which we refer to as \usik, and observed the effect of incrementally applying the series of different modelled effects (the miscellaneous noise effects are always applied). The effect of the improved architecture as well as the additional RGB channels are  also shown in the bottom two rows. 
The evaluation is performed on the real DiLiGenT dataset. This can be seen in Figure~\ref{fig:augmentations_converge} and  Figure~\ref{fig:incremental_increase} as well as in Table~\ref{table:aug}. We observed that the performance improves monotonically for most objects at each step, as well as the average error across the whole dataset. 

 \begin{table}[t]
\setlength{\tabcolsep}{0.25pt} 
\resizebox{1.0\columnwidth}{!}{%
\begin{tabular}{ | l | c c c c c c c c c c| l | }
 \hline
Method & Ball & Bear & Buddha & Cat & Cow & Goblet & Harvest & Pot1 & Pot2 & Reading & AVG  \\ \hline
Baseline \cite{Woodham1980} & 4.1 & 8.4 & 14.9 & 8.4 & 25.6 & 18.5 & 30.6 & 8.9 & 14.7 & 19.8 & 15.39  \\ 
SPLINE-Net\cite{zheng2019spline} & 1.7 & 4.7 & 9.1 & 5.5 & 9.6 & 9.4 & 24.4 & 5.9 & 7.9 & 12.8 & 9.1   \\ 
ICML \cite{taniai2018neural} & 1.5 & 5.8 & 10.4 & 5.4 & 6.3 & 11.5 & 22.6 & 6.1 & 7.8 & 11.0 & 8.83   \\ Exemplars \cite{hui2016shape} & 1.3 & 5.6 & 8.5 & 4.9 & 8.2 & 7.6 & 15.8 & 5.2 & 6.4 & 12.1 & 7.55   \\ 
PS-FCN \cite{chen2020deep} & 2.7 & 4.8 & 6.2 & 7.7 & 7.2 & 7.5 & 7.8 & 10.9 & 6.7 & 12.4 & 7.4  \\ \hline 
 CNN-PS\cite{ikehata2018cnn}, K=1 & 2.7 & 4.5 & 8.6 & 5 & 8.2 & 7.1 & 14.2 & 5.9 & 6.3 & 13 & 7.55   \\ 
CNN-PS\cite{ikehata2018cnn}, K=10 & 2.2 & 4.1 & 7.9 & 4.6 & 8 & 7.3 & 14 & 5.4 & 6 & 12.6 & 7.21   \\ \hline
Inv. model \cite{wang2020non}\textcolor{red}{$^{**}$} & 1.8 & 4.1 & 6.1 & 4.7 & 6.3 & 7.2 & 13.3 & 6.5 & 6.5 & 10.1 & 6.65 \\ \hline
\usik ,~K=1  & 2.2 &  3.7 &  7.8 &  4.3 &  5.5 &  7.6 & 15.0 &  5.4 &  5.7 & 10.9 &  6.79  \\
\usik ,~K=10  & 2.0 &	3.4 &	7.6 & 4.2 &	5.2 &	6.8 &	14.2 &	4.9 &	5.3 &	10.4 &	6.39  \\ 
\usbest ,~K=1& 2.0 & 3.6 & 7.6 &4.4 & 4.7 & 6.9 & 13.1 & 5.1 & 5.1 & 10.3 & 6.28 \\ 
\usbest ,~K=10 & 2.0 & 3.5 &	7.6 &	4.3 &	4.7 &	6.7 &	13.3 &	4.9 &	5.0 &	9.8 &		6.17 \\ \hline
\multicolumn{12}{c}{10 Lights} \\
\hline
Baseline \cite{Woodham1980} & 4.4 & 9.1 & 15.6 & 9.0 & 26.4 & 19.6 & 31.3 & 9.5 & 15.4 & 20.2 & 16.04 \\
CNN-PS\cite{ikehata2018cnn} & 9.1 & 11.7 & 13.2 & 14.1 & 14.7 & 14.6 & 15.5 & 17.0 & 14.0 & 19.6 & 14.34 \\
SPLINE-Net \cite{zheng2019spline} & 5.0 & 6.0  &10.1 & 7.5 & 8.8 & 10.4 & 19.1 & 8.8 & 11.8 & 16.1 & 10.35 \\
Mimify \cite{li2019learning} &  4.0 &	8.7 &	11.4 &	6.7 &	10.2 &	10.5 &	17.3 &	7.3 &	9.7 &	14.4 & 10.02 \\ 
Inv. model \cite{wang2020non}\textcolor{red}{$^{**}$} & 2.3 & 5.2 & 7.1 & 5.6 & 7.5 & 8.8 & 15.3 & 7.1 & 8.2 & 10.9 & 7.79\\ \hline
\usbest  ,~K=1 & \StackC{2.8 \\ $\pm0.4$ } & \StackC{5.2 \\ $\pm 0.4$} & \StackC{9.6\\$\pm 0.3$ } & \StackC{6.6\\ $\pm 0.2$ } & \StackC{7.8\\ $\pm 0.5$ } & \StackC{10.3\\ $\pm 0.9$ } & \StackC{16.5\\ $\pm 0.6$ } & \StackC{7.4 \\$\pm  0.2$} & \StackC{8.1 \\ $\pm 0.7$ } & \StackC{13.5 \\$\pm 0.4$ } & \StackC{8.76 \\ $\pm0.3$ } \\
\usbest ,~K=10 & \StackC{2.5 \\$\pm 0.4$ } & \StackC{4.9 \\ $\pm 0.3$ } & \StackC{9.4 \\ $\pm 0.3$} & \StackC{6.3 \\ $\pm 0.2$ } & \StackC{7.2 \\ $\pm 0.4$ } & \StackC{9.7 \\$\pm 0.8$ } & \StackC{16.1 \\ $\pm 0.5$ } & \StackC{7.0 \\$\pm 0.2$ } & \StackC{7.7 \\$ \pm 0.6$ } & \StackC{13.1 \\$\pm 0.5$ } & \StackC{8.37 \\$\pm 0.3$ }  \\
\hline
\end{tabular}
} 
\vspace{0.05cm}
\caption{Quantitative comparison of the proposed method (both simplified \usik ~and full \usbest ) on the DiLiGenT benchmark \cite{diligentshi2016benchmark}. For our networks as well as for \cite{ikehata2018cnn} results using K=10  are also presented for completeness. The bottom portion of the table presents evaluation with 10 random lights (\textcolor{red}{$^{**}$}\cite{wang2020non} uses 9x9 pixels patches and a specific illumination constraint so the comparison is not fully fair). We perform the experiment 10 times and report mean and standard deviation of the each error over the 10 tries (also computing K=10) \label{tab:main}}
\end{table}

It is noted that after the inclusion of the ambient, shadow and saturation variation steps, \usik ~outperform CNN-PS ($6.79^o$ vs $7.21^o$, see Table~\ref{tab:main}) which is trained with globally rendered data. This can be explained by the following three reasons. Firstly, the synthetic training data of CNN-PS do not include some of the effect we are modeling here namely light source brightness variation, noise and ambient light. Secondly, CNN-PS was trained on a limited subset of Disney material parameters (due to being constrained by slow, global object rendering). Finally, it is likely, that CNN-PS did overfit on the specific distribution of global effect of its training data rendered using only 15 meshes.

The superiority of our training data compared the ones of CNN-PS is also confirmed on the synthetic, globally rendered, objects of Figure~\ref{fig:blen} as well as the Cycles-PS dataset. \usbest ~outperforms CNN-PS in  of these 4 objects. The most significant improvement is on the PUMPKIN object ($5.33^o$ vs $9.89^o$) and it can be explained by the fact that CNN-PS does not include the discontinuity modelling which is important as the surface of this object is rough and detailed. The comparison its much closer on Cycles-PS ($11.36^o$ vs $11.8^o$) and this is expected as it is made in a similar way to the training data of Cycles-PS. The reason for the superiority of \usbest ~is probably the higher learning capacity and the inclusion of the RGB channels.  

\begin{figure}[t]
  \begin{minipage}{1.0\columnwidth} 
    
 \begin{sideways} {\scriptsize \hspace{0.5cm} Cow} \end{sideways} 
 \includegraphics[width=0.30\textwidth,trim={6.5cm 6cm 6.5cm  5cm},clip]{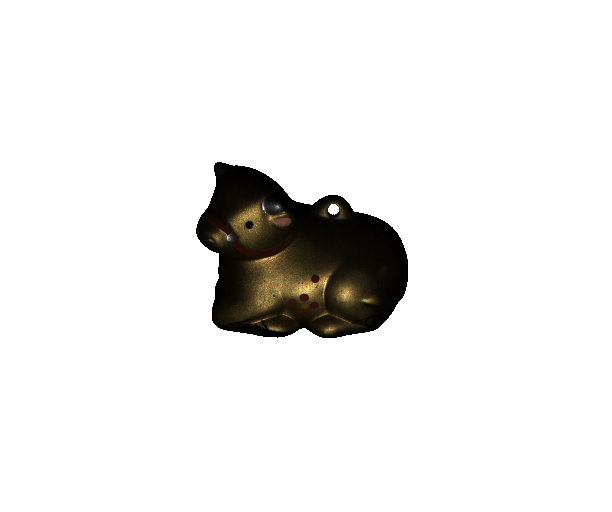} 
 \includegraphics[width=0.30\textwidth,trim={6.5cm 6cm 6.5cm  5cm},clip]{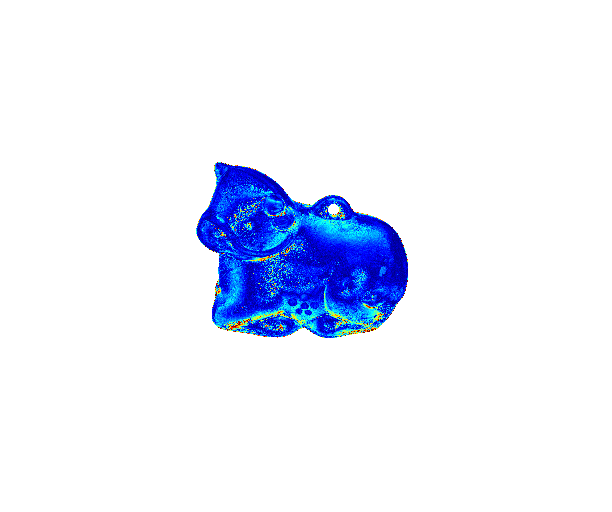} 
\includegraphics[width=0.30\textwidth,trim={6.5cm 6cm 6.5cm  5cm},clip]{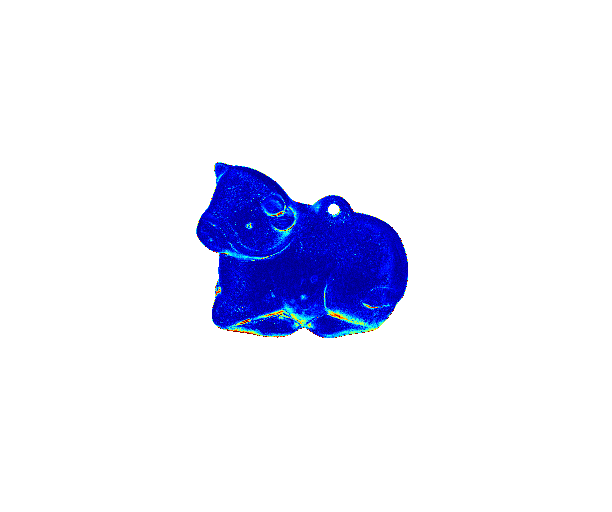} \\
\begin{sideways} {\scriptsize \hspace{0.5cm} Pot2} \end{sideways} 
\includegraphics[width=0.30\textwidth,trim={6.15cm 6cm 4cm  4cm},clip]{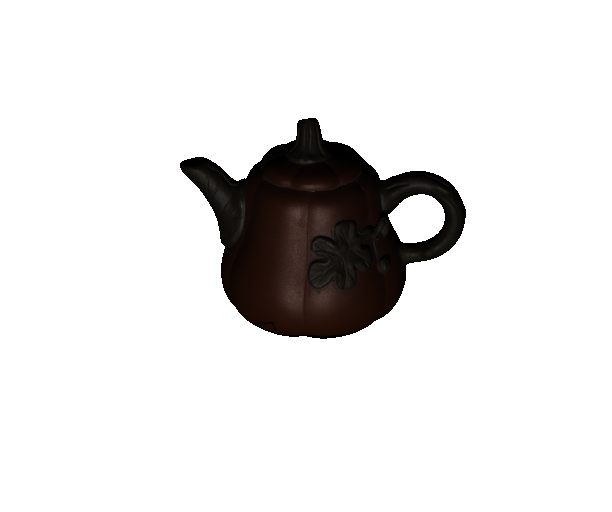} 
\includegraphics[width=0.30\textwidth,trim={6.15cm 6cm 4cm  4cm},clip]{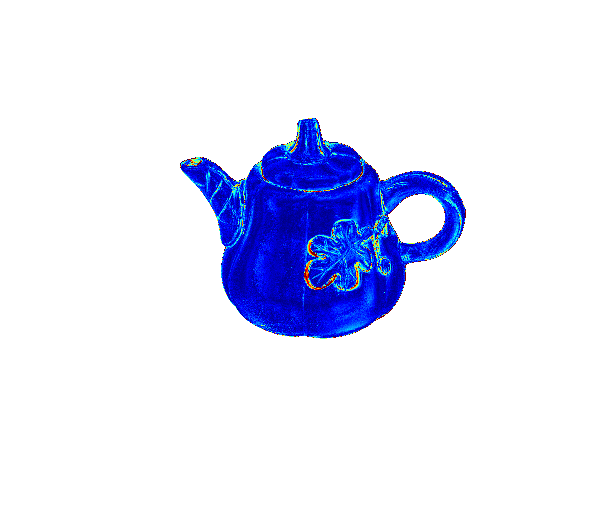}
\includegraphics[width=0.30\textwidth,trim={6.15cm 6cm 4cm  4cm},clip]{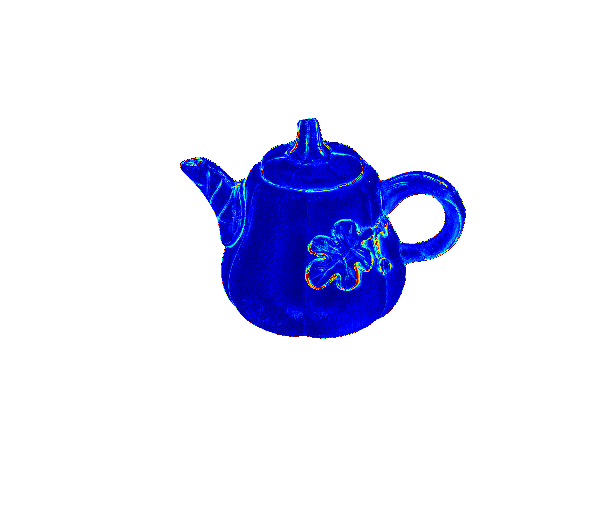} \\
\begin{sideways} {\scriptsize \hspace{0.25cm} Harvest} \end{sideways} 
\StackC{\includegraphics[width=0.30\textwidth,trim={4cm 6cm 4cm  4cm},clip]{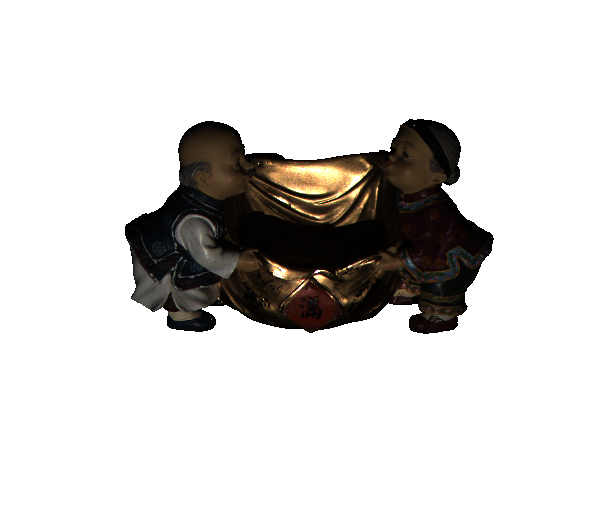} \\ \scriptsize Sample Image }
\StackC{\includegraphics[width=0.30\textwidth,trim={4cm 6cm 4cm  4cm},clip]{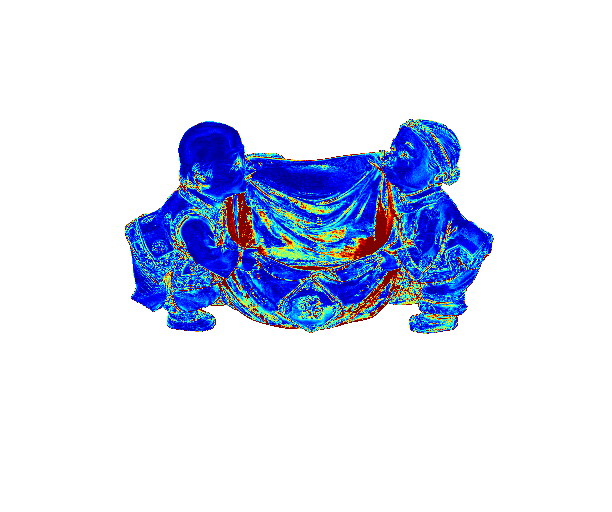} \\ \scriptsize CNN-PS \cite{ikehata2018cnn} }
\StackC{\includegraphics[width=0.30\textwidth,trim={4cm 6cm 4cm  4cm},clip]{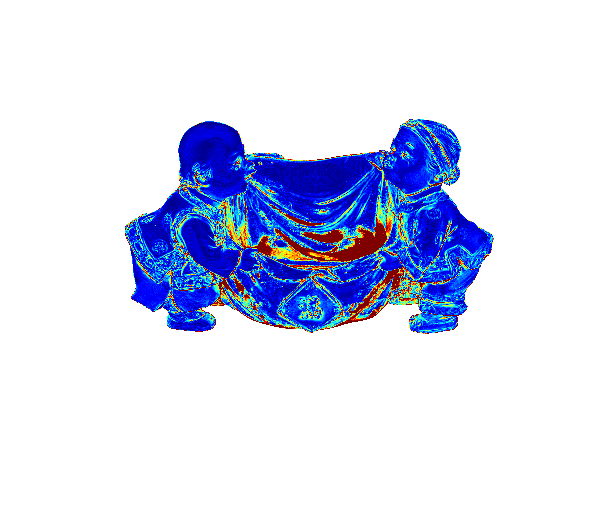} \\ \scriptsize Proposed \usbest} 
 \end{minipage}\begin{minipage}{0.09\textwidth}
    \hspace{-0.5cm}
\includegraphics[width=0.4\textwidth,trim={12.5cm 1.0cm 1.0cm  0.5cm},clip]{./images/colorbar_up.pdf}
  \end{minipage}
  \vspace{-0cm}
   \caption{Some error maps from Table~\ref{tab:main} (for k=1) comparing our \usbest ~to CNN-PS \cite{ikehata2018cnn}. It is noted that we significantly outperform the competition in convex region due to using more broad set of materials at train time (this is evident on the COW which is coated with metallic paint that is quite different that the metallic materials considered by CNN-PS). The Pot2 error map demonstrates the strength of our discontinuity augmentation on the leaf boundary. Finally, the Harvest error map shows that we can outperform in some concave regions (bellow left head); however regions with complicated  self-reflection patterns (middle of the image) are a potential limitation to our approach (due to the single bounce self reflection assumed.}
 \label{fig:mainqualitative}
 \end{figure}
 
\noindent
\textbf{Evaluation on MERL spheres.} The next synthetic experiment (Figure~\ref{fig:mrl}) compares both of our networks \usik ~(with all effects) and \usbest ~ with CNN-PS on the synthetic images rendered with MERL materials \cite{Matusik2003}. The aim of this evaluation is to demonstrate that our networks can deal with various real world reflectances. It is noted that \usik ~ outperforms CNN-PS ($6.1^o$ vs $4.3^o$ MAE) and this is expected as we include MERL materials in the training data. 
\usbest~further reduces the MAE to $4.8^o$, but there are still materials which are not very well recovered and thus this  motivates for future research.


\noindent
\textbf{Increasing size of globally rendered objects for CNN-PS.} In order to justify the use of our per-pixel training procedure, we examine the effect of increased amount of objects in the Cycles-PS~\cite{ikehata2018cnn} dataset to CNN-PS~\cite{ikehata2018cnn} network. We supplement original 15 objects of CNN-PS~\cite{ikehata2018cnn} with objects from \cite{diligentshi2016benchmark} Thingi10K~\cite{Thingi10K} dataset (see supplementary material for full list and examples of the objects). Test-time accuracy evolution of CNN-PS~\cite{ikehata2018cnn} network when trained in total on 20, 30, 40 and 50 objects is shown in Figure~\ref{fig:mrl} (right), using dashed lines. Note that all networks seem to achieve their best performance in early epochs and do not benefit from longer training. Also, adding more than 30 objects does not seem to benefit the performance on the DiLiGenT~\cite{diligentshi2016benchmark} dataset. This is likely due to overfitting to both global shapes of the objects provided in the dataset and particular choice of mixed-material rendering by ~\cite{ikehata2018cnn}. If some modelling of realistic effects (in this case, applied on globally rendered images) mentioned in Section~\ref{sec:Method} are applied such as ambient light, camera noise and multiplicative noise, the performance in initial epochs (solid lines) slightly improves to corresponding networks trained without data augmentation, however still the best performance of $7.11^{\circ}$ error is achieved by a network trained on 30 objects with no augmentations. In contrast PX-CNN-PS is able to achieve accuracy of $6.79^{\circ}$ by avoiding computationally inefficient global object rendering. 

\noindent
\textbf{Comparison with the state-of-the-art.} Finally, we compare our two networks with other state-of-the-art methods in the   DiLiGenT~\cite{diligentshi2016benchmark} dataset in Table~\ref{tab:main}, in both dense and sparse light settings. For completeness, we also include the results after applying the test time rotation pseudo-invariance augmentation (K=10). 
Three sample error maps are shown in Figure~\ref{fig:mainqualitative} (for the K=1 network evaluation). Both of our networks significantly outperform  the competition in the average error as well in almost all objects individually. The success of our method can be attributed on the ability of the network to deal with real world materials with complex reflectance (we exhibit very minimal error in convex regions where the PS problem reduced to BRDF inversion) as well as simultaneously being very robust to global illumination effects due to our modeling strategy. We note that the best performing method on the sparse setting \cite{wang2020non} uses 9x9 pixel patches as well as a constraint on the light setup at both train and test time (instead of 10 fully random lights) so the comparison is not fully fair (we still outperform it in the dense setting non the less).



\vspace{-0.2cm}
\section{Conclusion}
\label{sec:conclusion}
\vspace{-0.2cm}
In this work we presented a novel, simple and efficient concept for generating in-line training data for solving the PS problem, using a simple pixel observation map generation procedure. We approximate global effects like shadows, self-reflections, etc. by adopting a modeling strategy based on real and synthetic data observations. We analysed the performance of our approach while progressively adjusting the training data and we quantitatively showed the actual benefits in adopting such a modeling strategy. State-of-the-art results are achieved on the real DiLiGenT \cite{diligentshi2016benchmark} benchmark as well as the synthetic Cycles-PS \cite{ikehata2018cnn} one. 

Future work includes considering extension to mutli-view PS setting, e.g. using SDF representation \cite{logothetis2019differential}.



{\small
\bibliographystyle{ieee_fullname}
\bibliography{egbib}
}
\clearpage

\appendix

\section{Appendix}

This appendix provides supplementary material for the main document. 
 Section~\ref{sec:global} provides additional information about the 50 objects dataset used to train CNN-PS~\cite{ikehata2018cnn}. Section~\ref{sec:augm} provides an in depth explanation of the pixelwise data generation with all the relevant hyperparameters. Section~\ref{sec:res} contains additional visualisations of the qualitative results on the DiLiGent dataset.

 \section{CNN-PS Training on an Extended Dataset}
\label{sec:global}

\begin{figure}[t]
\centering
\includegraphics[height=0.085\textwidth,trim={0.0cm 0cm 0.0cm  0.0cm},clip]{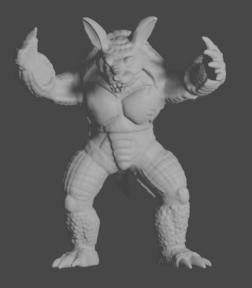} 
\includegraphics[height=0.085\textwidth,trim={0.0cm 0cm 0.0cm  0.0cm},clip]{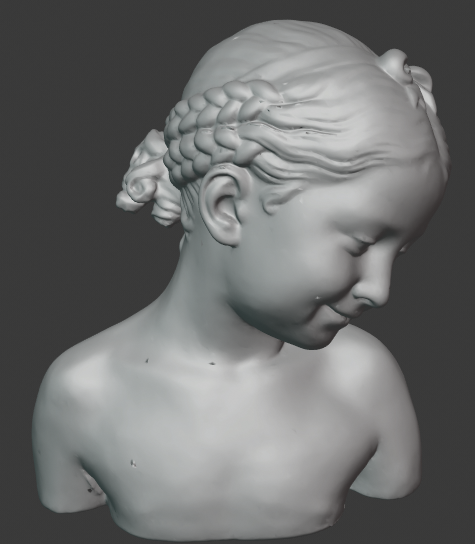} 
\includegraphics[height=0.085\textwidth,trim={0.0cm 0cm 0.0cm  0.0cm},clip]{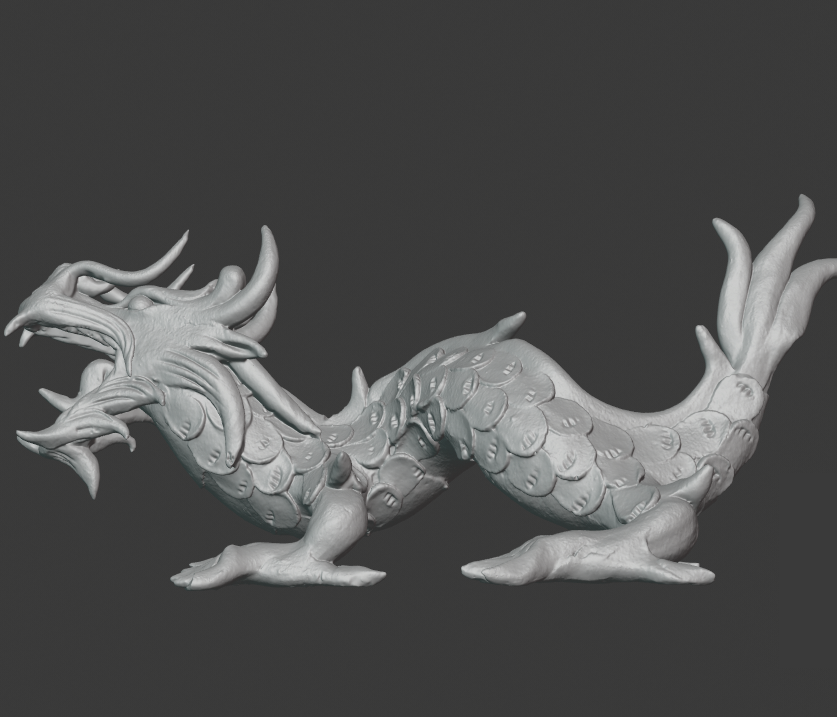} 
\includegraphics[height=0.085\textwidth,trim={0.0cm 0cm 0.0cm  0.0cm},clip]{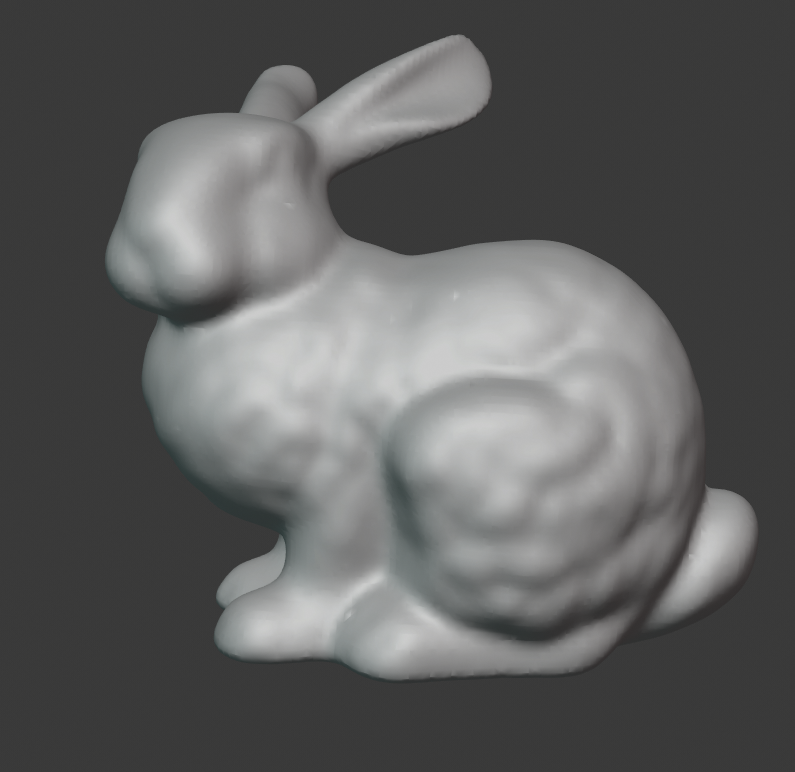} 
\includegraphics[height=0.085\textwidth,trim={0.0cm 0cm 0.0cm  0.0cm},clip]{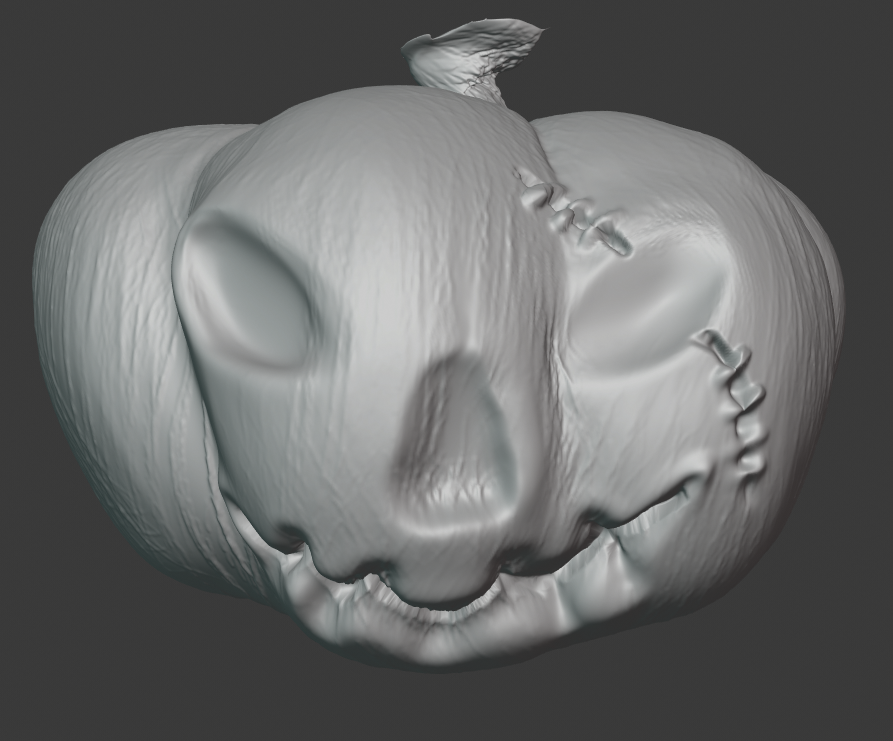} 
Additional objects for the 20 objects experiment \\
\includegraphics[height=0.085\textwidth,trim={0.0cm 0cm 0.0cm  0.0cm},clip]{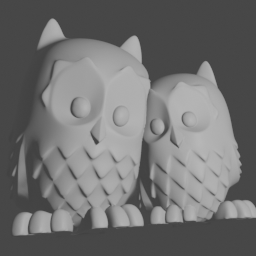} 
\includegraphics[height=0.085\textwidth,trim={0.0cm 0cm 0.0cm  0.0cm},clip]{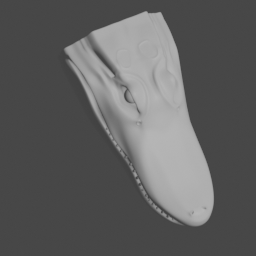} 
\includegraphics[height=0.085\textwidth,trim={0.0cm 0cm 0.0cm  0.0cm},clip]{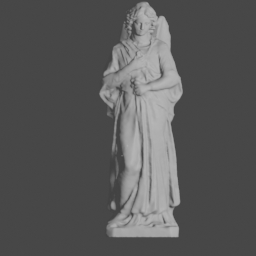} 
\includegraphics[height=0.085\textwidth,trim={0.0cm 0cm 0.0cm  0.0cm},clip]{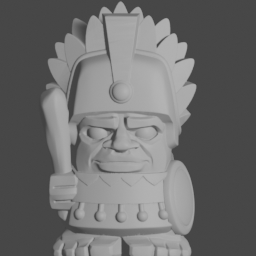} 
\includegraphics[height=0.085\textwidth,trim={0.0cm 0cm 0.0cm  0.0cm},clip]{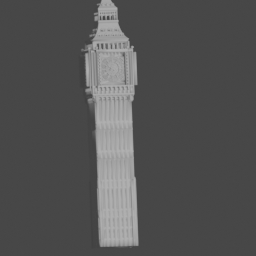} 
\includegraphics[height=0.085\textwidth,trim={0.0cm 0cm 0.0cm  0.0cm},clip]{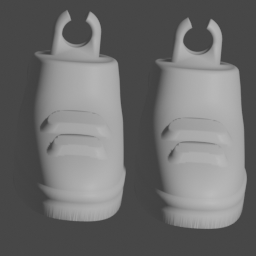} 
\includegraphics[height=0.085\textwidth,trim={0.0cm 0cm 0.0cm  0.0cm},clip]{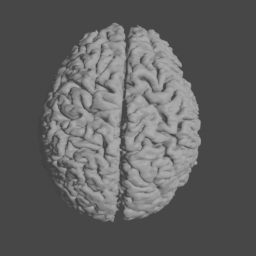} 
\includegraphics[height=0.085\textwidth,trim={0.0cm 0cm 0.0cm  0.0cm},clip]{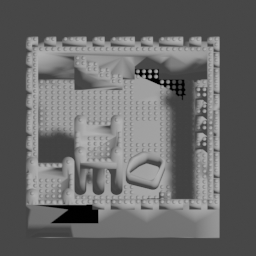} 
\includegraphics[height=0.085\textwidth,trim={0.0cm 0cm 0.0cm  0.0cm},clip]{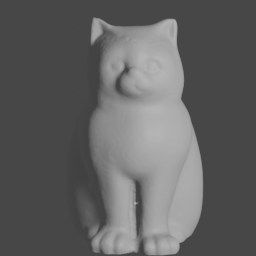} 
\includegraphics[height=0.085\textwidth,trim={0.0cm 0cm 0.0cm  0.0cm},clip]{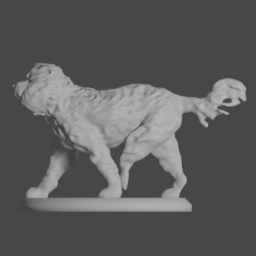} 
Additional objects for the 30 objects experiment \\
\includegraphics[height=0.085\textwidth,trim={0.0cm 0cm 0.0cm  0.0cm},clip]{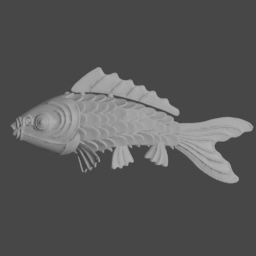} 
\includegraphics[height=0.085\textwidth,trim={0.0cm 0cm 0.0cm  0.0cm},clip]{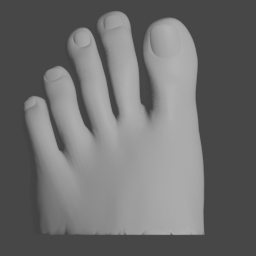} 
\includegraphics[height=0.085\textwidth,trim={0.0cm 0cm 0.0cm  0.0cm},clip]{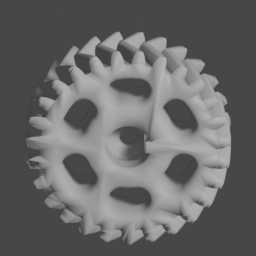} 
\includegraphics[height=0.085\textwidth,trim={0.0cm 0cm 0.0cm  0.0cm},clip]{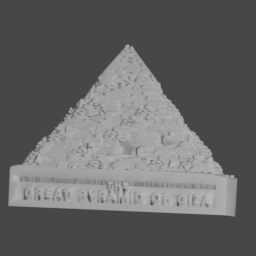} 
\includegraphics[height=0.085\textwidth,trim={0.0cm 0cm 0.0cm  0.0cm},clip]{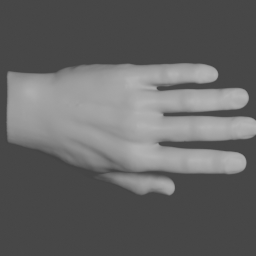} 
\includegraphics[height=0.085\textwidth,trim={0.0cm 0cm 0.0cm  0.0cm},clip]{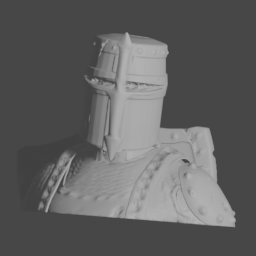} 
\includegraphics[height=0.085\textwidth,trim={0.0cm 0cm 0.0cm  0.0cm},clip]{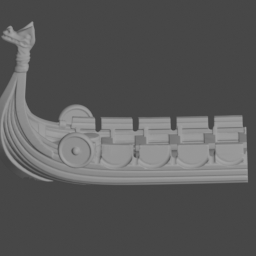} 
\includegraphics[height=0.085\textwidth,trim={0.0cm 0cm 0.0cm  0.0cm},clip]{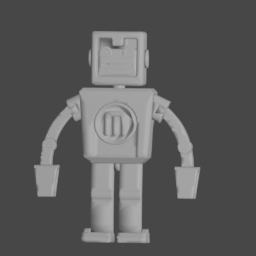} 
\includegraphics[height=0.085\textwidth,trim={0.0cm 0cm 0.0cm  0.0cm},clip]{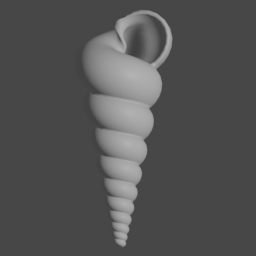} 
\includegraphics[height=0.085\textwidth,trim={0.0cm 0cm 0.0cm  0.0cm},clip]{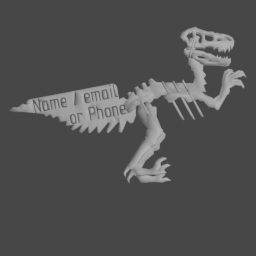} 
Additional objects for the 40 objects experiment \\
\includegraphics[height=0.085\textwidth,trim={0.0cm 0cm 0.0cm  0.0cm},clip]{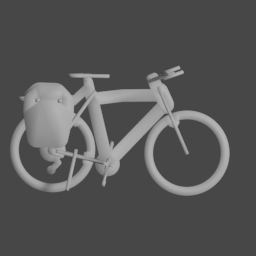} 
\includegraphics[height=0.085\textwidth,trim={0.0cm 0cm 0.0cm  0.0cm},clip]{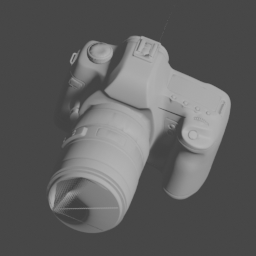} 
\includegraphics[height=0.085\textwidth,trim={0.0cm 0cm 0.0cm  0.0cm},clip]{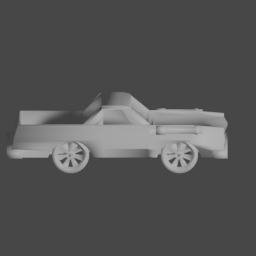} 
\includegraphics[height=0.085\textwidth,trim={0.0cm 0cm 0.0cm  0.0cm},clip]{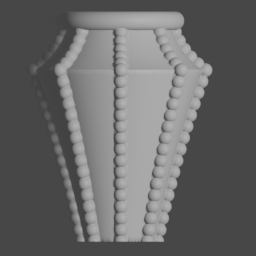} 
\includegraphics[height=0.085\textwidth,trim={0.0cm 0cm 0.0cm  0.0cm},clip]{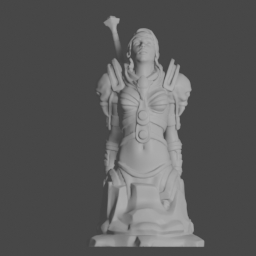} 
\includegraphics[height=0.085\textwidth,trim={0.0cm 0cm 0.0cm  0.0cm},clip]{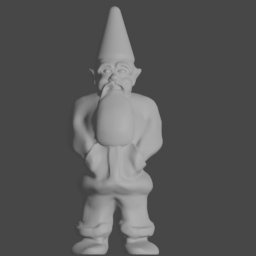} 
\includegraphics[height=0.085\textwidth,trim={0.0cm 0cm 0.0cm  0.0cm},clip]{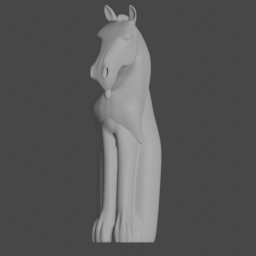} 
\includegraphics[height=0.085\textwidth,trim={0.0cm 0cm 0.0cm  0.0cm},clip]{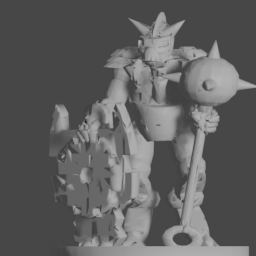} 
\includegraphics[height=0.085\textwidth,trim={0.0cm 0cm 0.0cm  0.0cm},clip]{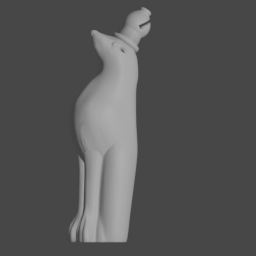} 
\includegraphics[height=0.085\textwidth,trim={0.0cm 0cm 0.0cm  0.0cm},clip]{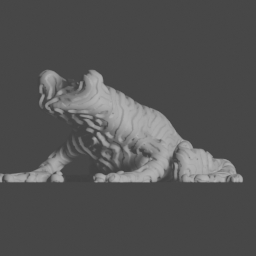} 
Additional objects for the 50 objects experiment \vspace{0.4cm}
 \caption{ Objects used to extend the training portion of the original Cycles-PS~\cite{ikehata2018cnn} dataset. } 
  \label{figlobalobjects}

\end{figure}

\begin{figure}[t]
\resizebox{1.0\columnwidth}{!}{\hspace{-0.08cm}
\begin{tabular}{ | c | c c c c c|}
\hline
Category & Roughness & Specular & Metallic & Subsurface & Other\\
\hline 
Diffuse &  $0.75-1$ & $0-0.25$ & $0$     & $0$ & $0-1$ \\
Specular&  $0-0.25$ & $0.75-1$ & $0-1$   & $0$ & $0-1$ \\
Metallic&  $0-1$  &    $0$     & $0.75-1$& $0$ & $0-1$ \\
\hline
\end{tabular}} \\
\includegraphics[width=0.1556\textwidth,trim={0.0cm 0cm 0.0cm  0.0cm},clip]{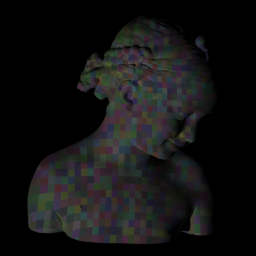} 
\includegraphics[width=0.1556\textwidth,trim={0.0cm 0cm 0.0cm  0.0cm},clip]{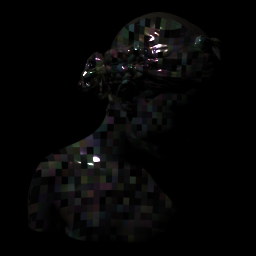} 
\includegraphics[width=0.1556\textwidth,trim={0.0cm 0cm 0.0cm  0.0cm},clip]{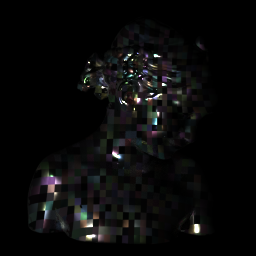} 
 \caption{Top: Material categories  considered for the extension of the CNN-PS~\cite{ikehata2018cnn} training set. Parameters are uniformly sampled in the respective ranges. \textit{Other} refers to all over parameters of the Disney BRDF namely \textit{BaseColor, SpecularTint, Sheen, SheeTint, Clearcoat, ClearcoatRoughness, IOR}. Bottom: sample images from these categories i.e. Diffuse (\textit{left}), specular dielectric (\textit{middle}) and metallic (\textit{right}). } 
 \label{figlobalmats}
\end{figure}

 This section provides additional information about the experiment during which CNN-PS~\cite{ikehata2018cnn} was trained on an extended dataset of globally rendered objects. This experiment is described in Section 5 and Figure 7 of the main publication. The dataset contains 50 objects in total, 15 of which are the original objects of the training portion of the Cycles-PS~\cite{ikehata2018cnn} dataset. These 15 objects are then supplemented with another 35 objects from Thingi10K~\cite{Thingi10K} dataset. CNN-PS is trained under four different setups, using 20, 30, 40 and 50 objects respectively. Figure~\ref{figlobalobjects} shows the images of the objects used for corresponding experiments. For each object, original rendering protocol of Cycles-PS~\cite{ikehata2018cnn} is employed to render 3000 images using Blender. 3 material categories (diffuse, specular dieletric and metallic) and 1000 random directional lights (uniform in the upper hemisphere up to $70^o$ elevation angle) per material category are sampled. The material hyper-parameter ranges are chosen to be slightly more general than the ones used in \cite{ikehata2018cnn}.  They are provided in Figure~\ref{figlobalmats}. All images are rendered at 256x256 resolution with each 8x8 pixel patch having a different material (i.e. 32x32 material maps). 

Training is performed using the original CNN-PS network architecture and training script of \cite{ikehata2018cnn} (using the same masking of lights per map and rotation augmentation procedure) with one change -  the Euclidean distance loss function is replaced with the angular error loss (see main submission, Section 4) to be fully comparable with our PX-CNN-PS.  We also performed an additional set of experiments (with 20, 30, 40, 50 objects), that also included generating additive and multiplicative noise as well as additive ambient light on top of obtained global renderings. These effects were implemented at train time as data augmentation (so as to get a different random sampling at each epoch) with values explained in Table~\ref{table:augm}.



\section{Pixelwise Data Generation}
\label{sec:augm}

This section provides a more in depth explanation of the pixelwise data generating procedure. The objective is to approximate certain global illumination effects as well as other real world imperfections so as the generated observation maps to  be realistic enough to be applicable in real world test data. It is important to note that we do not  attempt to exactly replicate the distribution of many real world effects, but instead avoid significantly underestimating them. Thus most of the approximations are aimed to be the reasonable upper bound of an effect and in practice they are likely to be much less (e.g. the sampled ambient effect is up to $0.01$, but for most real pixels it is likely to be a lot less). This is a valid procedure as long as the network has enough learning capacity.

As a general principle, we aim to sample most parameters uniformly in the appropriate range, in order to avoid data bias. To simplify the notation, we assume that the pixel value $i$ is a  real number $i \in [0,1]$  with $0$ being completely black and $1$ being the saturation level (although the division with the light source brightness can lead to observation map values higher than 1). We also denote a uniform real distribution in the interval $[a,b]$ as $\mathcal{U_R}(a,b)$, a uniform integer one in the interval $[k,l]$ as $\mathcal{U_I}(k,l)$ and a normal distribution with mean $\mu$ and standard deviation $\sigma$ as $\mathcal{N}(\mu,\sigma)$.

\paragraph{\textbf{Normals:}} We sample normals uniformly in the upper hemisphere so as to maximise the generality of the training data. Note that for some very oblique normals, after all the effects are applied, all of the map pixels may be ending up very small. To avoid numerical instabilities, any map where the maximum RGB pixel value is less than 1e-3 (i.e. 0.1\% of the saturation level) is discarded and not included in the training data.

\paragraph{\textbf{Lights:}} We trained 2 different networks, aimed to tackle the dense and sparse light settings. For the dense lights settings, we matched the distribution of \cite{ikehata2018cnn} which was 50-1000 random lights (up to $70^o$ elevation angle).  The sparse light setup was made to match that of \cite{li2019learning} namely  exactly 10 random lights with elevation angle of up to $45^o$.  For both dense and sparse light settings, light source brightness $\phi$ is sampled uniformly and independently (for all lights and channels) in the DiLiGenT range, i.e. $\phi=\mathcal{U_R}(0.28,3.2)$.



\paragraph{\textbf{Materials:}} The material determines the surface albedo (which is essentially the intrinsic color) as well as other BRDF parameters. We sample albedo $\rho$ color components uniformly so $\rho_{red}=\mathcal{U_R}(0,1)$ and similarly for green and blue channels.  In order for our data to be applicable in a range of real world situations, 75\%  of the data are generated  using a random material from the Disney BRDF.  All 8 parameters (excluding \textit{subsurface},\textit{IOR}), namely \textit{metallic,  specular, roughness, specularTint, sheen,  sheenTint, clearcoat} and \textit{clearcoatRoughness} are sampled uniformly ($\mathcal{U_R}(0,1)$) and independently. The Dinsey non-linear equation uses both the albedo and BRDF parameters as inputs so the direct reflectance $r_d$ component computation is straightforward. The remaining 25\% of the training data are generated using data from the MERL material database. Unfortunately, this database only contains 100 materials with specific albedo (e.g. ``blue-acrylic'', ``green-latex'') which is a very limited set for tackling general PS problems. To overcome this limitation, we generate virtual materials superset of the MERL database with the following procedure: firstly a random MERL material $M \sim \mathcal{U_I}(1,100)$ is selected. Then a random weight $w \sim \mathcal{U_R}(0,1)$ is sampled. Finally, the material's BRDF $B_M$ is mixed with a Lambertian component $\vect{N} \cdot \vect{L}$ to get overall reflectance:

\begin{equation}
r_{b}(\vect{N}, \vect{L}, \vect{V}_0,\rho,M)=\rho \Big(w B_M(\vect{N}, \vect{L}, \vect{V}_0) +(1-w)\vect{N} \cdot \vect{L} \Big )    
\label{eq:merl}
\end{equation}
where in Equation~\ref{eq:merl},  $\vect{L}, \vect{V}_0$ are the light and view direction respectively (as explained in the main text).

We note that the above material sampling procedure is aimed to target general test data. Of course, in the case of a very specific application (e.g industrial inspection), a more constrained set of materials would be more appropriate.


\begin{figure}[t]

\includegraphics[width=0.45\textwidth,trim={0.0cm 0cm 0.0cm  0.5cm},clip]{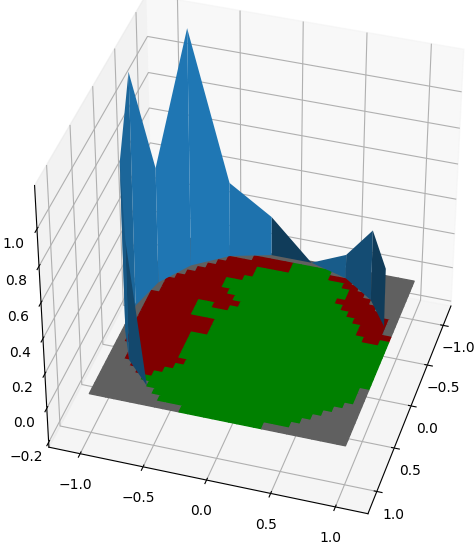} 
 \caption{ Demonstration of the assumed shadow model consisting of a circular ``wall'' surrounding the observation map. All shaded direction are marked red whereas non shaded ones are marked green. }
 \label{fig:shadow}
\end{figure}
\paragraph{\textbf{Cast shadows:}} Cast shadows are observed in real data when a part of the surface is blocking the light, thus turning the direct reflectance to zero. Our aim is to compute a shadow map approximation, (i.e. compute all of the directions were direct refletance is blocked) regardless of which light sources are actually available.  
We note that that realistic  shadows are likely to be piece-wise continuous in scenes with a few discrete objects.  In addition, the likelihood of a particular directing to be in shadow increases as its elevation angle increases, and in principle, the central direction $\vect{V}_0=[0,0,1]$ is never blocked by shadow. Finally, if a direction is blocked by shadow, it is very likely that all other directions with the same azimuth and more oblique elevation angles to also be blocked by shadow.

Taking all of the above considerations into account, the assumed shadow model consists of a circular ``wall'' (see Figure\ref{fig:shadow}) surrounding the observation map. More specifically, we sample 20 height values (corresponding to azimuth angles which are integer multiples of $36^o$) from a Gaussian with mean 0, standard deviation 2 (taking the absolute value). In addition, to allow for the possibility of no shadow in a range of directions, each of these values is set to 0 with a chance of 25\%. We also sample 25\% of the data with completely empty shadow maps which are aimed to approximate points is convex part of a surface.  Then, the height of the ``wall'' is linearly interpolated in order to get a height value for all azimuth angles. Finally, each pixel of the shadow map is set as $0$ if the corresponding direction (extending out of the center) is intersecting the wall.



\begin{table}[t]
\begin{center}
\resizebox{1.0\columnwidth}{!}{%
\begin{tabular}{ | l | c  c  c  c  c  c  c  c  c  c  | c | }
\hline
Effects & Ball & Bear & Buddha & Cat & Cow & Goblet & Harvest & Pot1 & Pot2 & Reading & Mean  \\ \hline
Up to Ambient       & 2.08 &  4.24 &  8.38 &  4.59 &  5.77 &  7.79 & 14.88 &  5.57 &  6.01 & 12.46 &  7.18 \\ 
$+$Same mat. reflection    & 2.40 &  3.70 &  7.97 &  4.46 &  5.76 &  7.58 & 14.51 &  5.37 &  5.62 & 10.91 &  6.83 \\
$+$Discontinuity & 2.17 &  3.72 &  7.78 &  4.25 &  5.49 &  7.56 & 14.95 &  5.36 &  5.69 & 10.94 &  6.79 \\ \hline
$+$Diff. mat. reflection   &  1.99 & 	3.98 & 	8.59 & 	4.65 & 	6.59 & 	8.29 & 	15.01 & 	5.59 & 	6.70 & 	11.24 & 7.26 \\
$+$Discontinuity &2.56 &	3.67 &	8.13 &	4.51 &	6.48 &	7.69 &	14.72 &	5.60 &	5.95 &	11.74  &7.10 \\ \hline
\end{tabular}} \vspace{0.1cm}
\caption{Justification for computing self reflection with reflecting points having the same material (but different albedo than the main point). The 3 lines on top are copied from Table 1 of main text whereas the 2 lines on bottom contain the self reflection effect computing with different material for each self reflection point. These different materials were sampled but perturbing the main Diligent material parameters by a random value $\in [-0.1,0.1]$.  \label{table:ref}}\vspace{-0.4cm}
\end{center}
\end{table}

\paragraph{\textbf{Self reflections:}} As explained in the main text, we approximate the self reflection effect by sampling a few directions $\vect{L}_R$ (inside the shadow map) and then computing a single light bounce from  $\vect{L}$ to  $\vect{L}_R$ to  $\vect{V}_0$. Table \ref{table:ref} justifies using the same material for these reflecting points (which aims to model the case of an object with piece-wise constant material distribution). We note that in real objects, highly concave regions are most likely to have the maximum amount of self reflection which also corresponds to the maximum amount of shadows. To enforce this positive correlation between the self reflection magnitude and the amount of shadows, we first sample 5 directions $\vect{L}_R$ uniformly in the upper hemisphere, and then only keep those that are part of the shadow map, i.e. $S(\vect{L}_R)=0$. Thus, the more shaded pixels the shadow map has, the higher the chance for a higher number of self reflection points. Note that this sampling procedure only makes a subset of data points that have a non empty shadow map to have any self reflection directions; this is consistent with real data as significant self reflection is only present in a small portion of the data (corresponding to highly concave regions of surfaces or nearby very reflective objects).  

\paragraph{\textbf{Surface discontinuity:}} For this step, we allow 15\% of the training data to be a 
combination of 2 or 3 `subpixels', each having a different normal  $\vect{N}_k$ and albedo $\rho_k$. As explained in the main text, all direct reflection $r_d$ and self reflection $r_r$ components computed and averaged. We note that the overall normal (used for training the network) is simply the average of the subpixel's normals.


\paragraph{\textbf{Ambient light:}} Ambient light aims to address any additional reflection, such as from objects in the background or the even the atmosphere. Literature usually assumes a constant reflection for all light sources as the multiple bounces tend to average out the effect. None the less, as this reflectance component is caused by the light source, the ambient effect has to be proportional to its brightness $\phi$ . In addition, it is reasonable to assume high correlation with the surface albedo (ambient reflection at a dark point should be dark, ambient reflection at a red point should be red etc.) as well as diminished reflection at oblique angles so we assume correlation with $\vect{N} \cdot \vect{V}_0$). Therefore, we set $a= \rho \vect{N} \cdot \vect{V}_0  \mathcal{U_R}(0,0.01) $ and apply this effect to 75 \% of the generated data points (i.e. 25\% of out training data are made ambient free). We note that $a$ is the same for all light sources and its contribution to the total reflectance is multiplied by the light brightness $\phi$ (see Equation 5 in main text).

\begin{table}[t]
\begin{center}
\resizebox{0.99\columnwidth}{!}{%
\begin{tabular}{ | l |c | c | }
\hline
Effect & Probability & Magnitude\\
\hline
Number of Lights & N/A  &  $\mathcal{U_I}(50,1000)$ dense /10 sparse  \\
Light Brightness & N/A & $\mathcal{U_R}(0.28,3.2)$ \\
MERL Materials & 0.25  & N/A \\
Shadow & 0.75 &   $|\mathcal{N}(0,2)|$ \\
Self reflections & * & $\leq 5$ \\
Surface Discontinuity & 0.15 & $\mathcal{U_I}(2,3)$ \\
Ambient $a$ & 0.75 & $\rho \vect{N} \cdot \vect{V}_0  \mathcal{U_R}(0,0.01)$ \\
Noise Multiplicative & 1 & $\mathcal{U_R}(0.95,1.05) \mathcal{N}(1,10^{-4})$ \\
Noise Additive & 1 & $\mathcal{U_R}(-10^{-4},10^{-4}) + \mathcal{N}(0,10^{-4})$ \\
Quantisation & 1 & 16 bits \\
\hline
\end{tabular}
}\vspace{0.2cm}
\caption{Summary of all the data generation hyperparameters.  The probability column determines the proportion of the data with that particular effect doing applied (some effects like discretisation and noise are always used). the Note that self reflections are only sampled when the shadow map is non empty, and the corresponding magnitude is a positively correlated with the amount of shaded pixels. \label{table:augm}
}
\end{center}
\end{table}

\paragraph{\textbf{Noises:}} As explained in the main text, we apply four different types of noise namely additive and multiplicative, uniform and Gaussian. The most important component is the uniform multiplicative as it is aimed to address several unmodeled effects (i.e. near light attenuation) which effect pixel intensities multiplicatively, and thus was set to 5\%, i.e. $n_{MU}= \mathcal{U_R}(0.95,1.05)$. The rest of the hyper-parameters for Equation 5 were: multiplicative Gaussian noise $n_{MG}=\mathcal{N}(1,10^{-3})$, Gaussian additive noise $n_{AG}=\mathcal{N}(0,10^{-4})$ and uniform additive noise $n_{AU}= \mathcal{U_R}(-10^{-4},10^{-4})$. 


  

All of the relevant hyperparameters are summarised in Table~\ref{table:augm} and their application order is shown in Equation 5 in the main document . 

\subsection{Diligent Results}
\label{sec:res}

\begin{figure*}[ht]
\centering
\StackC{\begin{sideways} {CNN-PS, K=1} \end{sideways} \\ MAE}
 \StackC{\includegraphics[width=0.185\textwidth,trim={7.0cm 5cm 5.0cm  0.0cm},clip]{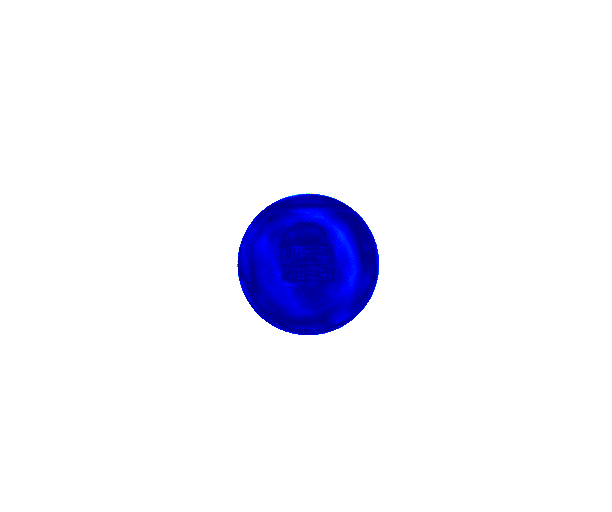} \\ $2.7^o$}
 \StackC{\includegraphics[width=0.185\textwidth,trim={7.0cm 5cm 5.0cm  0.0cm},clip]{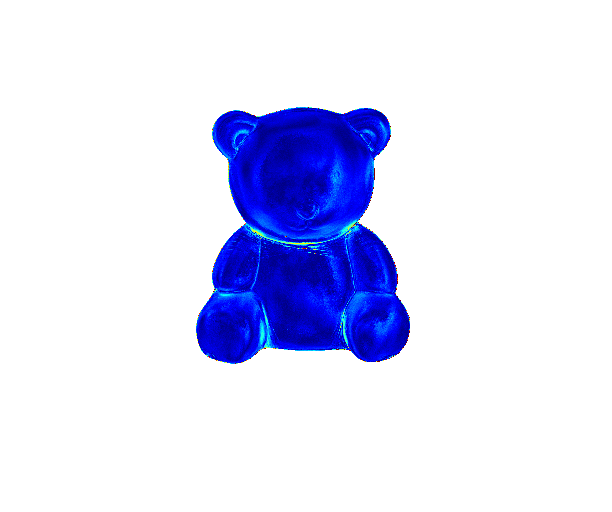} \\  $4.5^o$}
 \StackC{\includegraphics[width=0.185\textwidth,trim={7.0cm 5cm 5.0cm  0.5cm},clip]{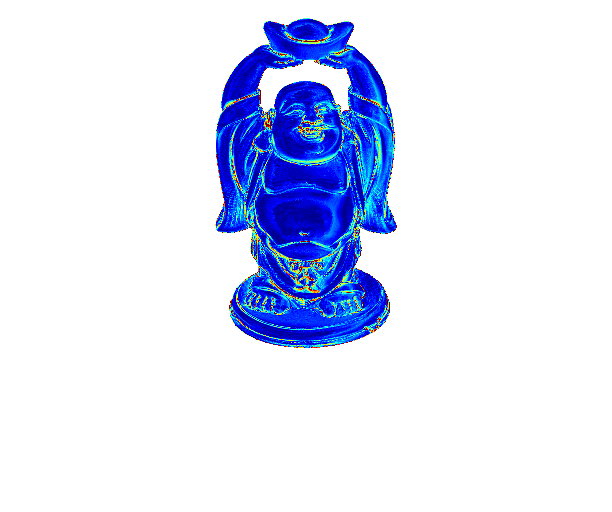} \\  $8.6^o$}
 \StackC{\includegraphics[width=0.185\textwidth,trim={7.5cm 5cm 4.5cm  0.0cm},clip]{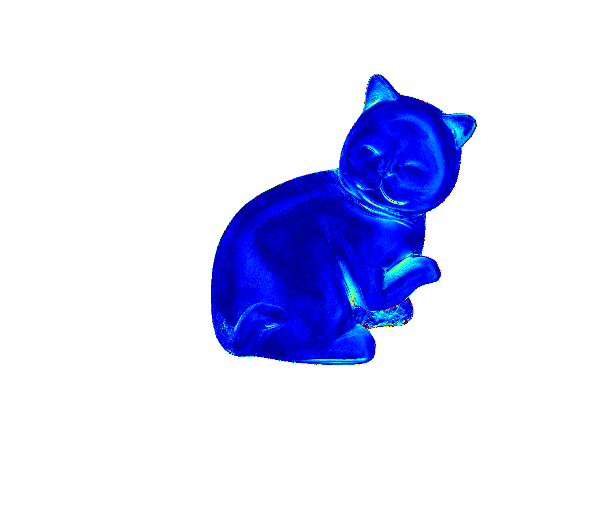} \\  $5^o$}
 \StackC{\includegraphics[width=0.185\textwidth,trim={7.0cm 5cm 5.0cm  0.0cm},clip]{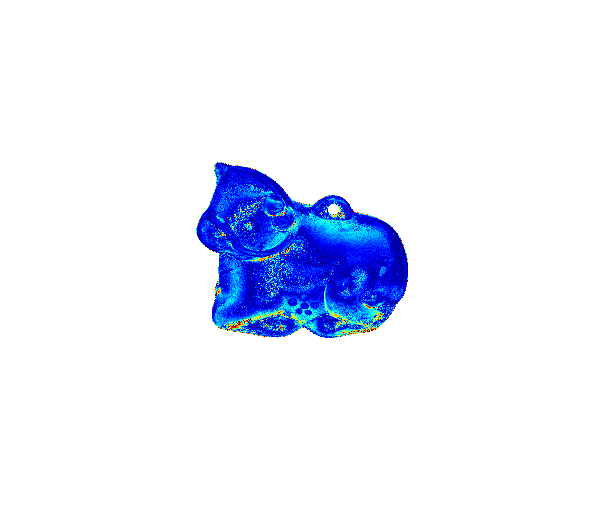} \\  $8.2^o$} \\
 
 \StackC{\begin{sideways} {PX-NET, K=1} \end{sideways} \\ MAE}
 \StackC{\includegraphics[width=0.185\textwidth,trim={7.0cm 5cm 5.0cm  0.0cm},clip]{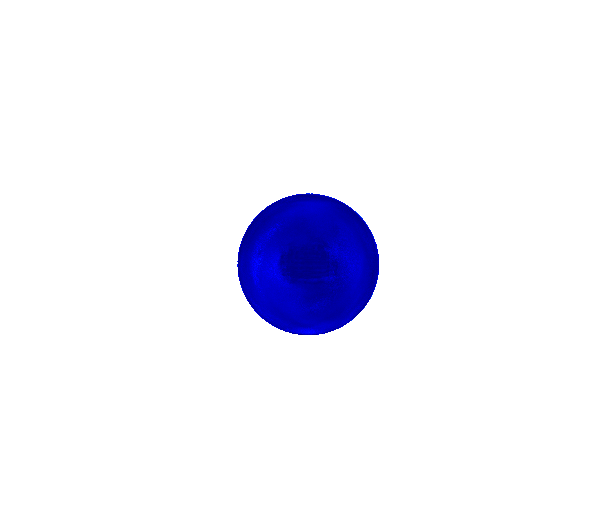} \\ $2.0^o$}
 \StackC{\includegraphics[width=0.185\textwidth,trim={7.0cm 5cm 5.0cm  0.0cm},clip]{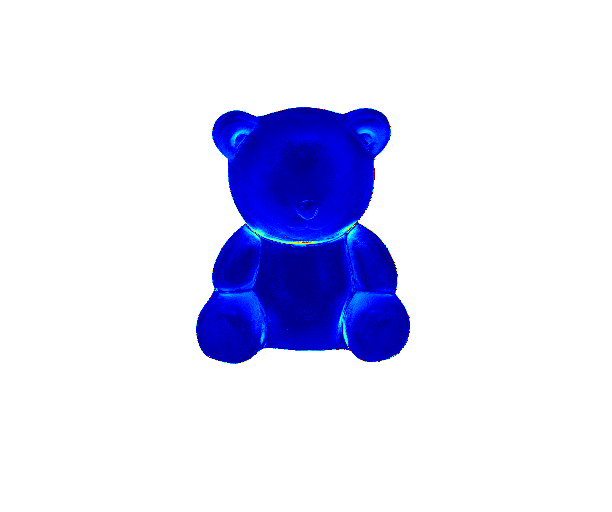} \\  $3.6^o$}
 \StackC{\includegraphics[width=0.185\textwidth,trim={7.0cm 5cm 5.0cm  0.5cm},clip]{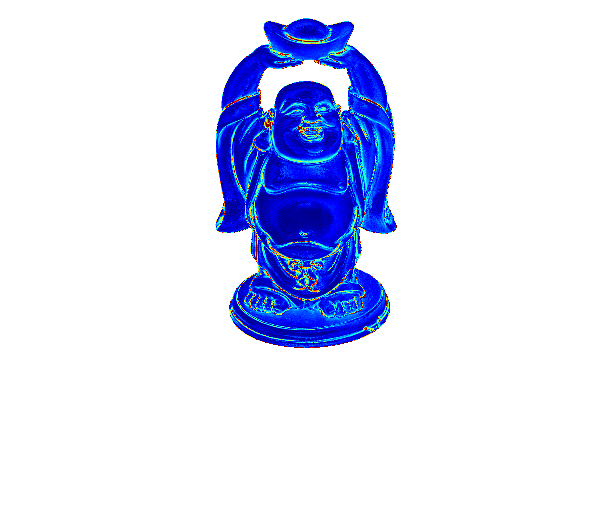} \\  $7.6^o$}
 \StackC{\includegraphics[width=0.185\textwidth,trim={7.5cm 5cm 4.5cm  0.0cm},clip]{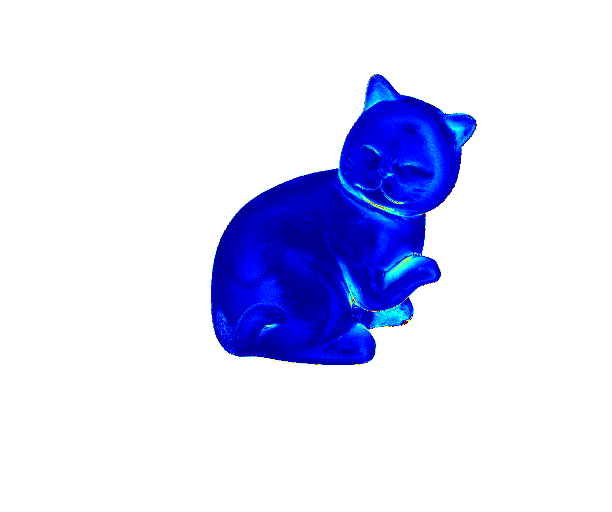} \\  $4.4^o$}
 \StackC{\includegraphics[width=0.185\textwidth,trim={7.0cm 5cm 5.0cm  0.0cm},clip]{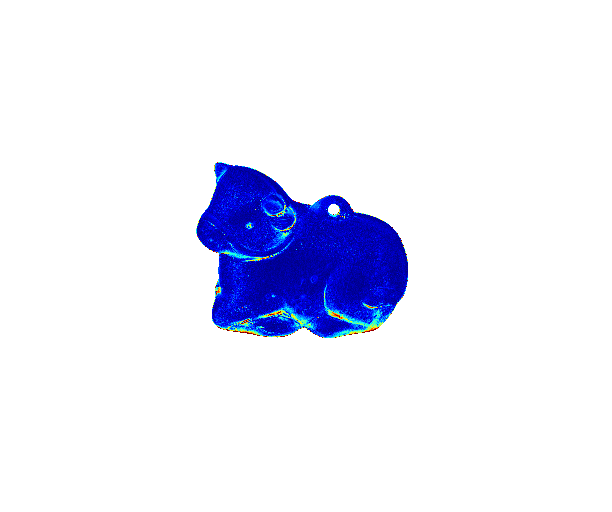} \\  $4.7^o$} \\
 
 \StackC{\begin{sideways} {CNN-PS, K=10} \end{sideways} \\ MAE}
 \StackC{\includegraphics[width=0.185\textwidth,trim={7.0cm 5cm 5.0cm  0.0cm},clip]{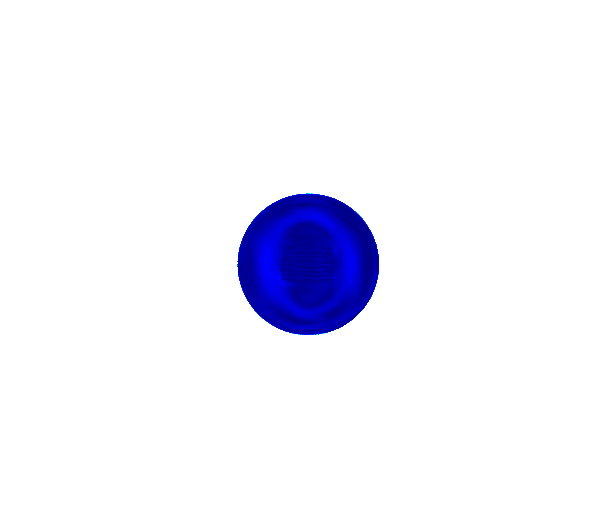} \\ $2.2^o$}
 \StackC{\includegraphics[width=0.185\textwidth,trim={7.0cm 5cm 5.0cm  0.0cm},clip]{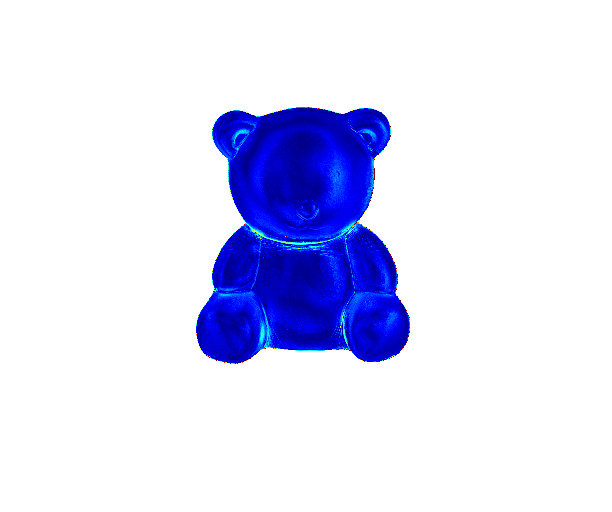} \\  $4.1^o$}
 \StackC{\includegraphics[width=0.185\textwidth,trim={7.0cm 5cm 5.0cm  0.5cm},clip]{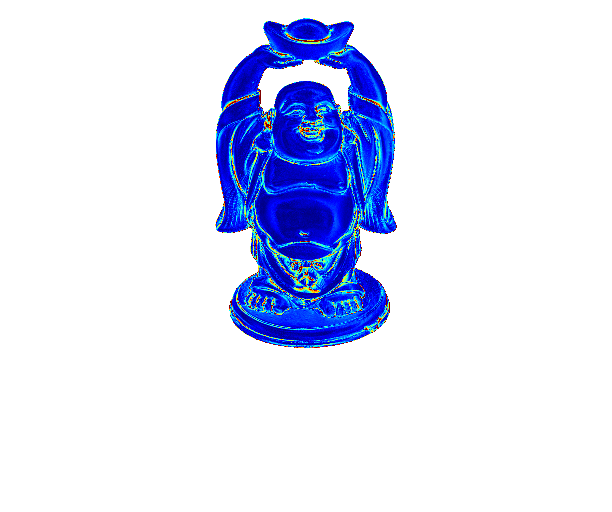} \\  $7.9^o$}
 \StackC{\includegraphics[width=0.185\textwidth,trim={7.5cm 5cm 4.5cm  0.0cm},clip]{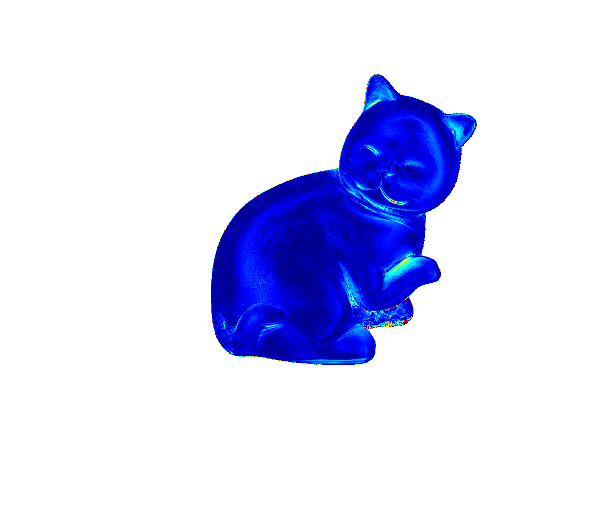} \\  $4.6^o$}
 \StackC{\includegraphics[width=0.185\textwidth,trim={7.0cm 5cm 5.0cm  0.0cm},clip]{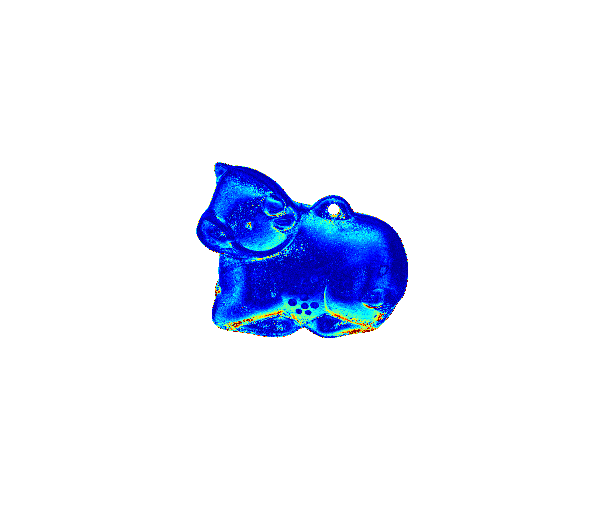} \\  $8.^o$} \\
 
 \StackC{\begin{sideways} {PX-NET, K=10} \end{sideways} \\ MAE}
 \StackC{\includegraphics[width=0.185\textwidth,trim={7.0cm 5cm 5.0cm  0.0cm},clip]{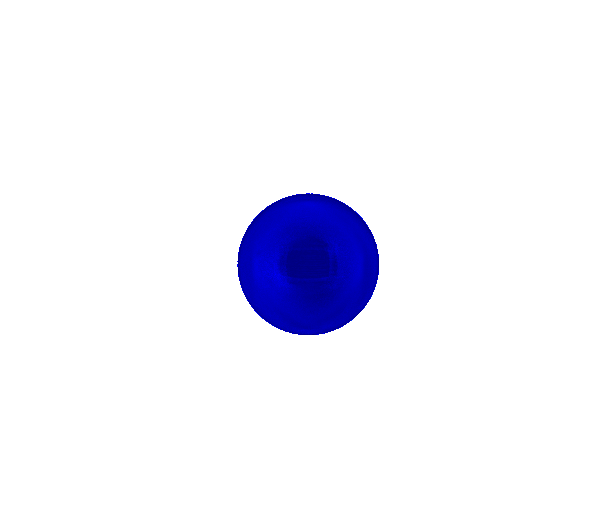} \\ $2.0^o$}
 \StackC{\includegraphics[width=0.185\textwidth,trim={7.0cm 5cm 5.0cm  0.0cm},clip]{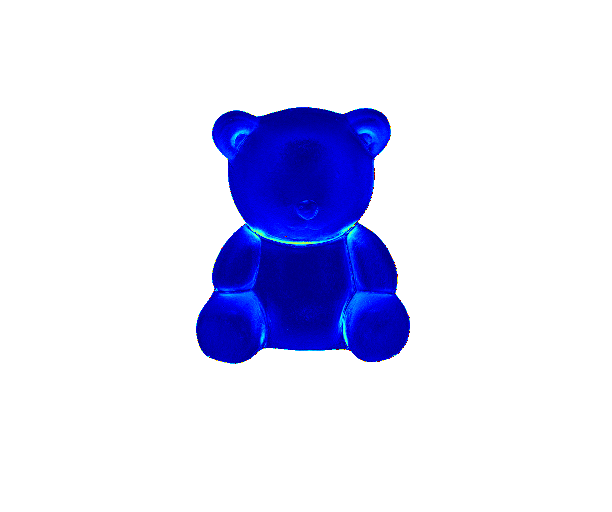} \\  $3.5^o$}
 \StackC{\includegraphics[width=0.185\textwidth,trim={7.0cm 5cm 5.0cm  0.5cm},clip]{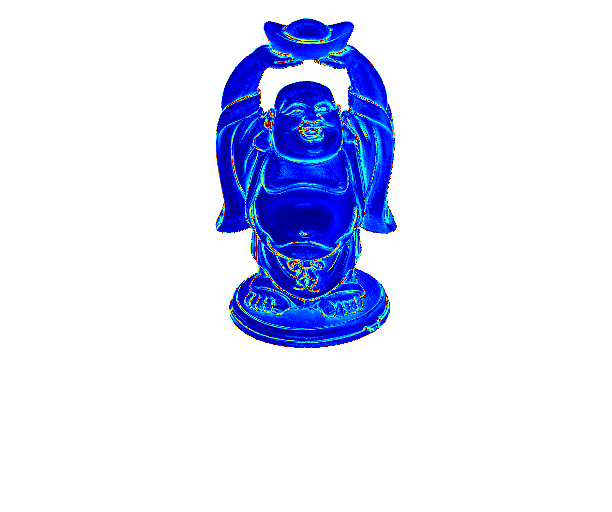} \\  $7.6^o$}
 \StackC{\includegraphics[width=0.185\textwidth,trim={7.5cm 5cm 4.5cm  0.0cm},clip]{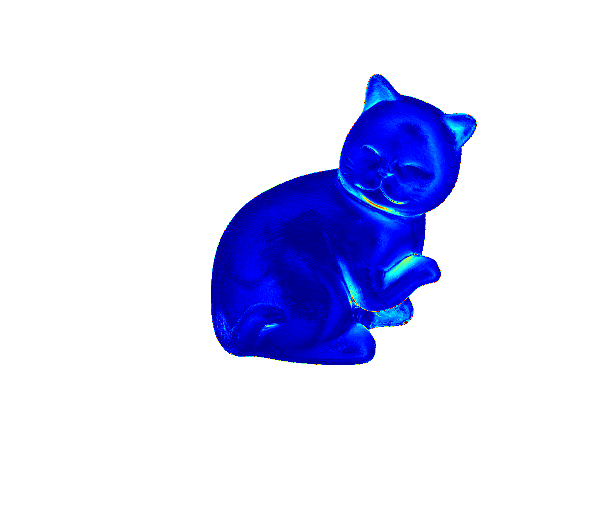} \\  $4.3^o$}
 \StackC{\includegraphics[width=0.185\textwidth,trim={7.0cm 5cm 5.0cm  0.0cm},clip]{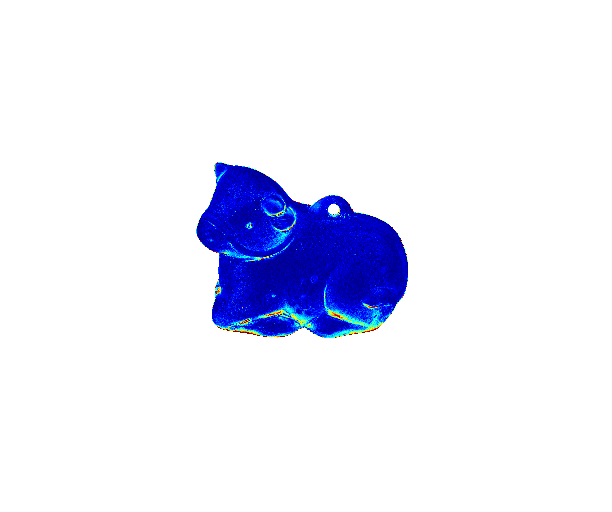} \\  $4.7^o$} \\
\includegraphics[width=1.0\textwidth,trim={0.0cm 1cm 0.0cm  8.5cm},clip]{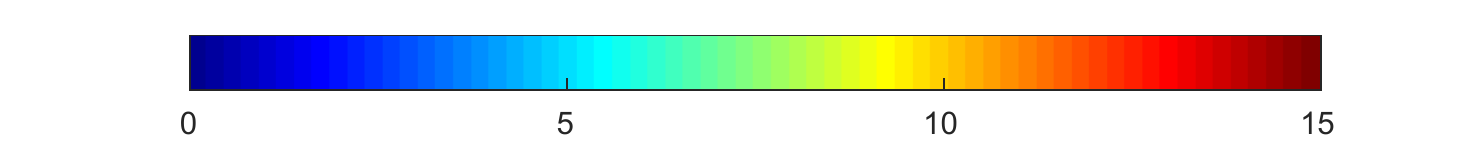}
 \caption{Visual comparison [1/2] of CNN-PS \cite{ikehata2018cnn} with the proposed PX-NET  (Table 2 of the main paper ) for both K=1 and K=10  on the Diligent dataset.} 
 \label{fig:res:real:all1}
\end{figure*}

\begin{figure*}[ht]
\centering
\StackC{\begin{sideways} {CNN-PS, K=1} \end{sideways} \\ MAE}
 \StackC{\includegraphics[width=0.185\textwidth,trim={5.0cm 5cm 5.0cm  1.0cm},clip]{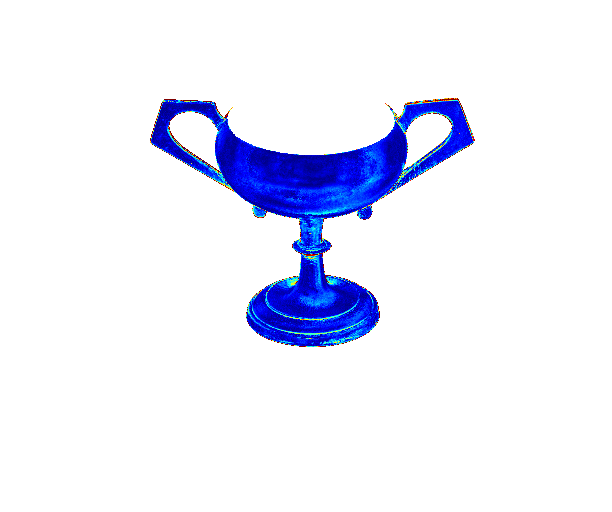} \\ $7.1^o$}
 \StackC{\includegraphics[width=0.185\textwidth,trim={4.0cm 5cm 4.0cm  1.0cm},clip]{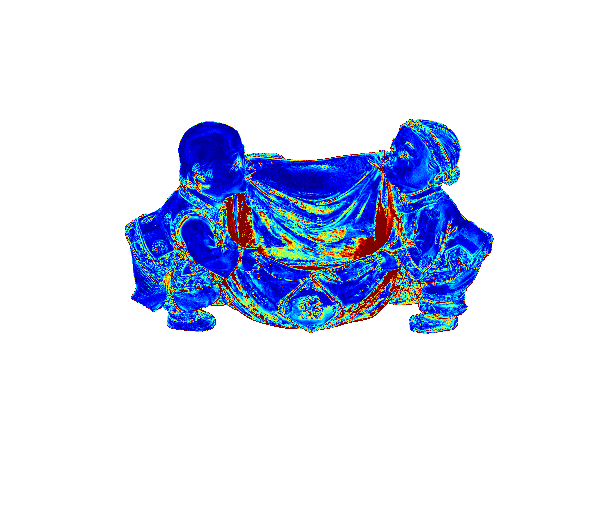} \\  $14.2^o$}
 \StackC{\includegraphics[width=0.185\textwidth,trim={4.0cm 5cm 3.0cm  1.5cm},clip]{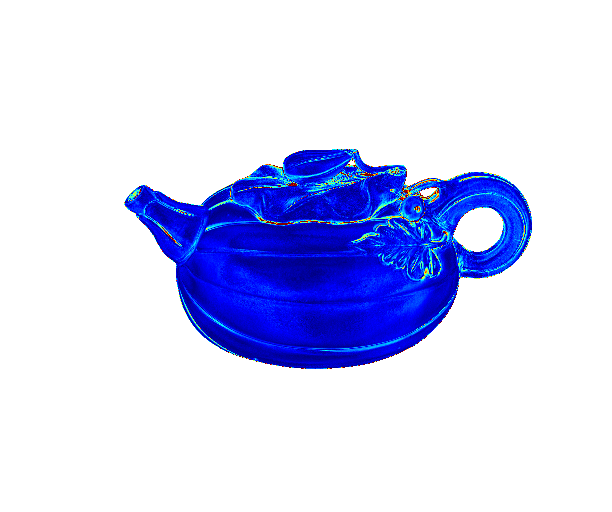} \\  $5.9^o$}
 \StackC{\includegraphics[width=0.185\textwidth,trim={6.5cm 5cm 4.5cm  1.0cm},clip]{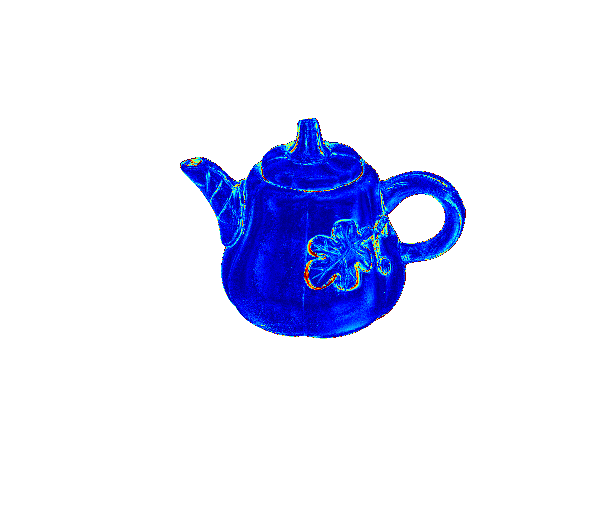} \\  $6.3^o$}
 \StackC{\includegraphics[width=0.185\textwidth,trim={7.0cm 5cm 5.0cm  1.0cm},clip]{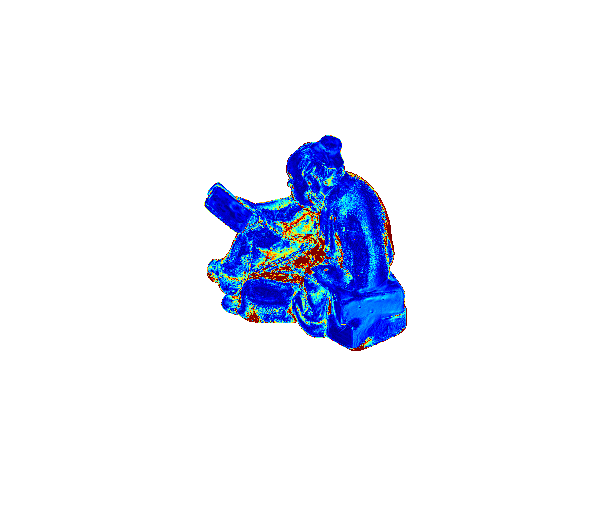} \\  $13^o$} \\
 
 \StackC{\begin{sideways} {PX-NET, K=1} \end{sideways} \\ MAE}
 \StackC{\includegraphics[width=0.185\textwidth,trim={5.0cm 5cm 5.0cm  1.0cm},clip]{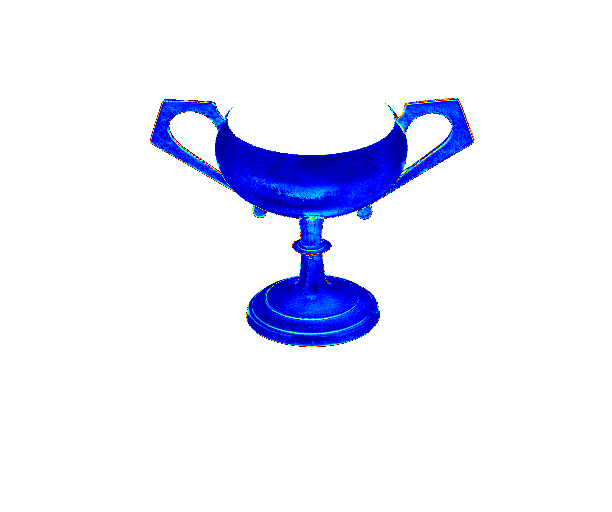} \\ $6.9^o$}
 \StackC{\includegraphics[width=0.185\textwidth,trim={4.0cm 5cm 4.0cm  1.0cm},clip]{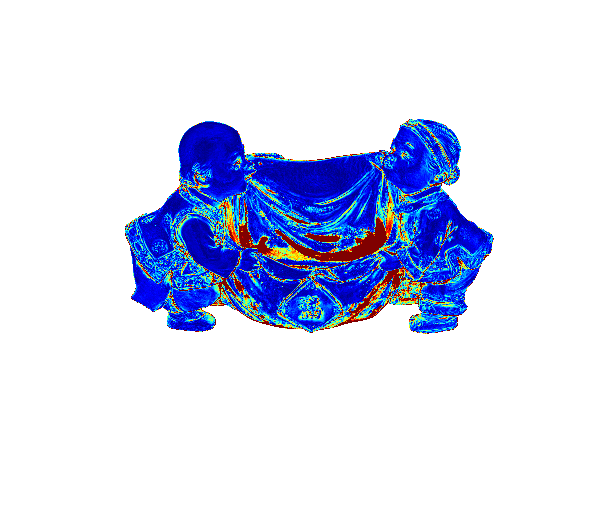} \\  $13.1^o$}
 \StackC{\includegraphics[width=0.185\textwidth,trim={4.0cm 5cm 3.0cm  1.5cm},clip]{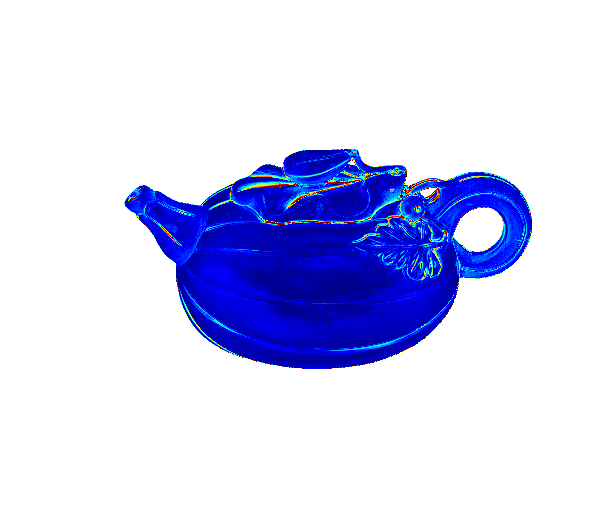} \\  $5.1^o$}
 \StackC{\includegraphics[width=0.185\textwidth,trim={6.5cm 5cm 4.5cm  1.0cm},clip]{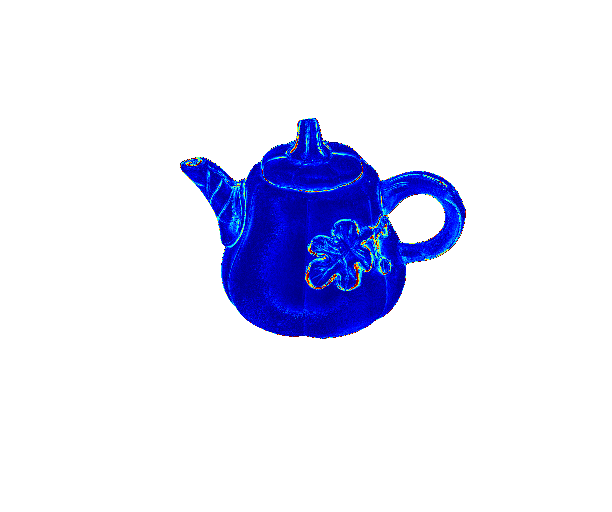} \\  $5.1^o$}
 \StackC{\includegraphics[width=0.185\textwidth,trim={7.0cm 5cm 5.0cm  1.0cm},clip]{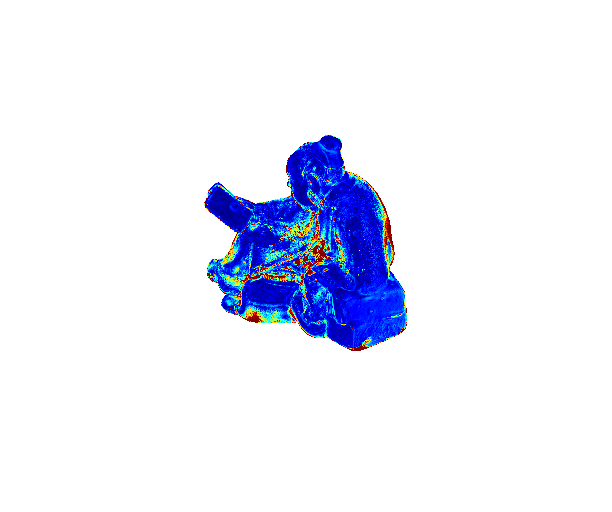} \\  $10.3^o$} \\
 
 \StackC{\begin{sideways} {CNN-PS, K=10} \end{sideways} \\ MAE}
 \StackC{\includegraphics[width=0.185\textwidth,trim={5.0cm 5cm 5.0cm  1.0cm},clip]{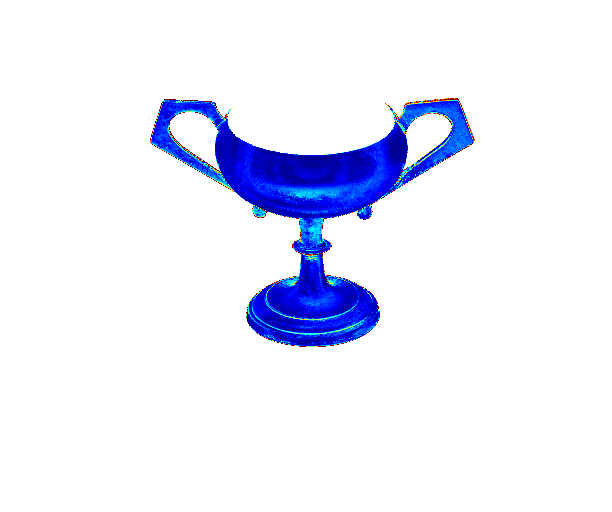} \\ $7.3^o$}
 \StackC{\includegraphics[width=0.185\textwidth,trim={4.0cm 5cm 4.0cm  1.0cm},clip]{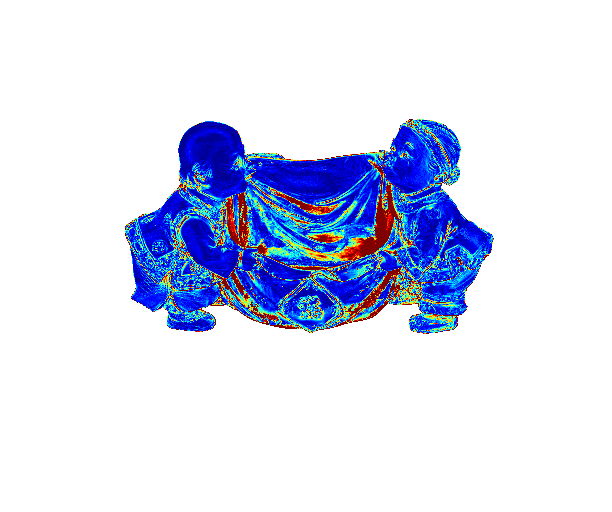} \\  $14^o$}
 \StackC{\includegraphics[width=0.185\textwidth,trim={4.0cm 5cm 3.0cm  1.5cm},clip]{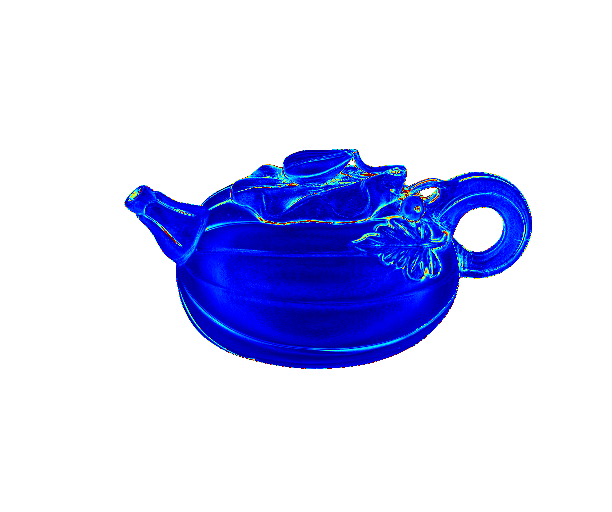} \\  $5.4^o$}
 \StackC{\includegraphics[width=0.185\textwidth,trim={6.5cm 5cm 4.5cm  1.0cm},clip]{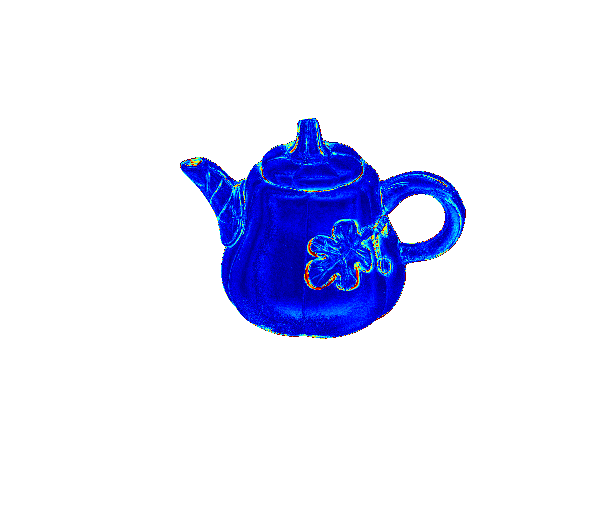} \\  $6^o$}
 \StackC{\includegraphics[width=0.185\textwidth,trim={7.0cm 5cm 5.0cm  1.0cm},clip]{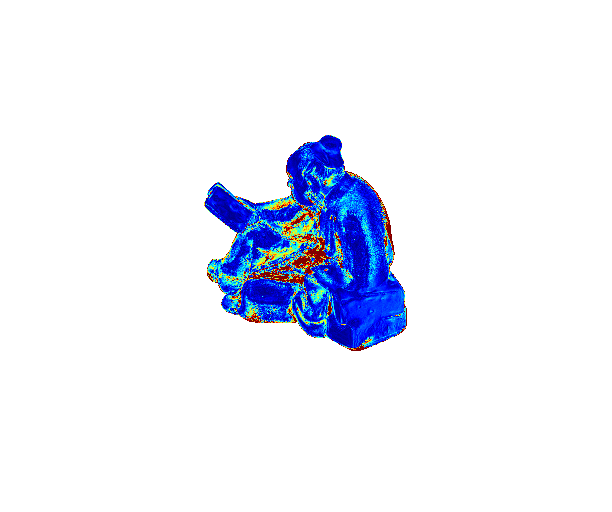} \\  $12.6^o$} \\
 
 \StackC{\begin{sideways} {PX-NET, K=10} \end{sideways} \\ MAE}
 \StackC{\includegraphics[width=0.185\textwidth,trim={5.0cm 5cm 5.0cm  1.0cm},clip]{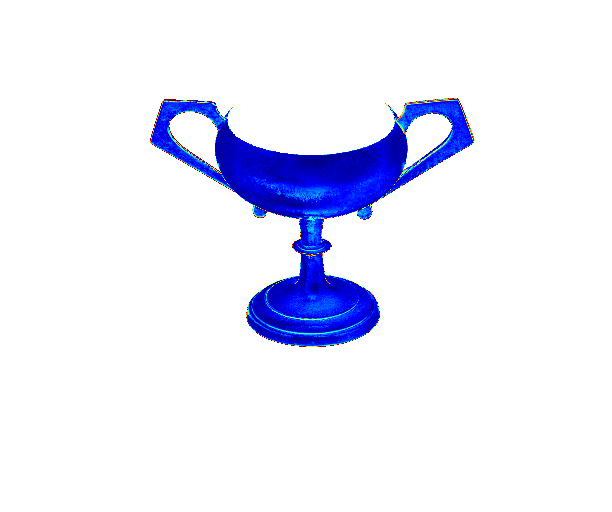} \\ $6.7^o$}
 \StackC{\includegraphics[width=0.185\textwidth,trim={4.0cm 5cm 4.0cm  1.0cm},clip]{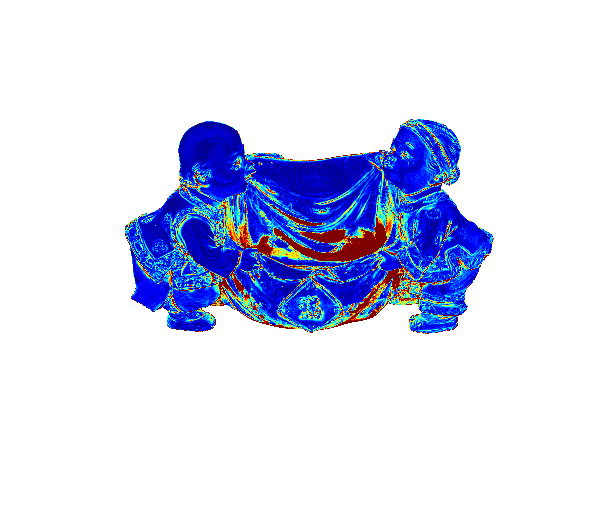} \\  $13.3^o$}
 \StackC{\includegraphics[width=0.185\textwidth,trim={4.0cm 5cm 3.0cm  1.5cm},clip]{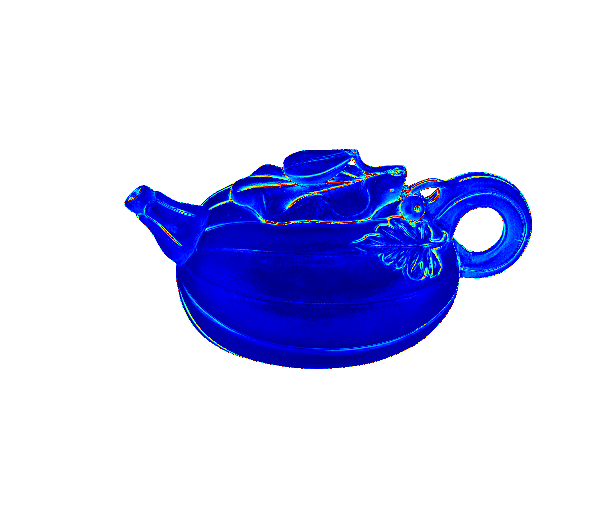} \\  $4.9^o$}
 \StackC{\includegraphics[width=0.185\textwidth,trim={6.5cm 5cm 4.5cm  1.0cm},clip]{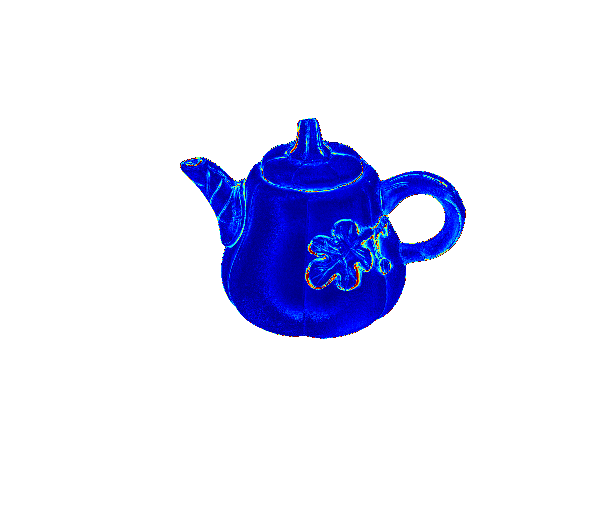} \\  $5.0^o$}
 \StackC{\includegraphics[width=0.185\textwidth,trim={7.0cm 5cm 5.0cm  1.0cm},clip]{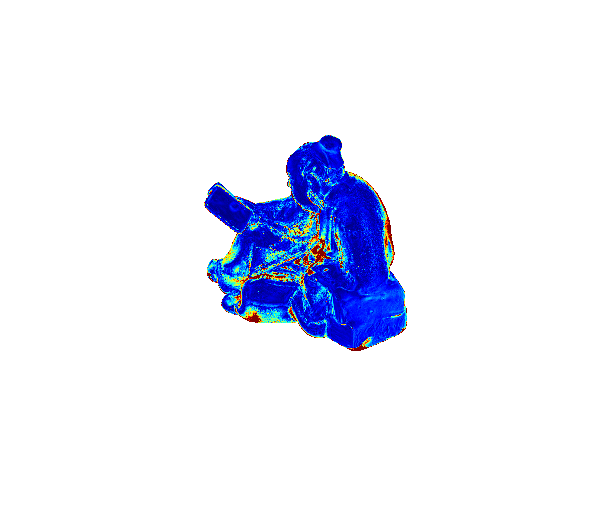} \\  $9.8^o$} \\
\includegraphics[width=1.0\textwidth,trim={0.0cm 1cm 0.0cm  8.5cm},clip]{./images/colorbar_down.pdf}

 \caption{Visual comparison [2/2] of CNN-PS \cite{ikehata2018cnn} with the proposed PX-NET  (Table 2 of the main paper ) for both K=1 and K=10  on the Diligent dataset.} 
 \label{fig:res:real:all2}
\end{figure*}

This section contains full visual comparison with CNN-PS \cite{ikehata2018cnn} for all Diligent objects at Figures~\ref{fig:res:real:all1} and \ref{fig:res:real:all2}.

\end{document}